\documentclass[10pt,twocolumn,letterpaper]{article}

\usepackage[pagenumbers]{cvpr} 

\usepackage{graphicx}
\usepackage{amsmath}
\usepackage{amssymb}
\usepackage{booktabs}
\usepackage{multirow}
\usepackage{makecell}
\usepackage[table,xcdraw]{xcolor}
\usepackage{bbm}

\usepackage[pagebackref,breaklinks,colorlinks]{hyperref}

\abovecaptionskip=\skip27  
\belowcaptionskip=\skip28  

\usepackage[capitalize]{cleveref}
\crefname{section}{Sec.}{Secs.}
\Crefname{section}{Section}{Sections}
\Crefname{table}{Table}{Tables}
\crefname{table}{Tab.}{Tabs.}

\newcommand{\tly}{}

\def\cvprPaperID{7264} 
\def\confName{CVPR}
\def\confYear{2022}

\begin{document}

\title{Contrastive Boundary Learning for Point Cloud Segmentation}

\author{
Liyao Tang$^1$, Yibing Zhan$^2$, Zhe Chen$^{1}$, Baosheng Yu$^1$, Dacheng Tao$^{2,1}$ \\
$^1$ The University of Sydney, Australia $^2$ JD Explore Academy, China \\
{\tt\small ltan9687@uni.sydney.edu.au, zhanyibing@jd.com, \{zhe.chen1, baosheng.yu\}@sydney.edu.au} \\
{\tt\small dacheng.tao@gmail.com}
}

\maketitle


\begin{abstract}
Point cloud segmentation is fundamental in understanding 3D environments. However, current 3D point cloud segmentation methods usually perform poorly on scene boundaries, which degenerates the overall segmentation performance. In this paper, we focus on the segmentation of scene boundaries. Accordingly, we first explore metrics to evaluate the segmentation performance on scene boundaries. To address the unsatisfactory performance on boundaries, we then propose a novel contrastive boundary learning (CBL) framework for point cloud segmentation. Specifically, the proposed CBL enhances feature discrimination between points across boundaries by contrasting their representations with the assistance of scene contexts at multiple scales.
By applying CBL on three different baseline methods, we experimentally show that CBL consistently improves different baselines and assists them to achieve compelling performance on boundaries, as well as the overall performance, \eg in mIoU.
The experimental results demonstrate the effectiveness of our method and the importance of boundaries for 3D point cloud segmentation.
Code and model will be made publicly available at \href{https://github.com/LiyaoTang/contrastBoundary}{https://github.com/LiyaoTang/contrastBoundary}.
\end{abstract}


\section{Introduction}
\label{sec:intro}

\begin{figure}
\begin{center} 
  \includegraphics[width=\linewidth]{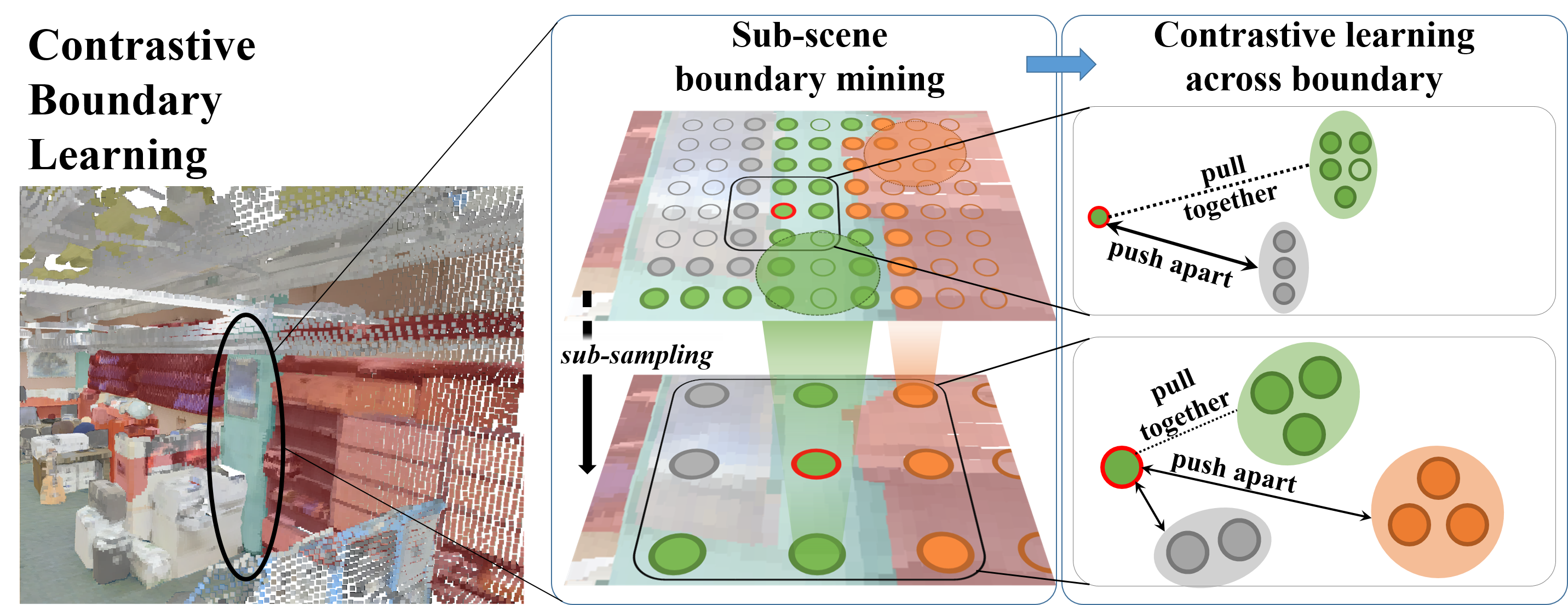}
   \includegraphics[width=\linewidth]{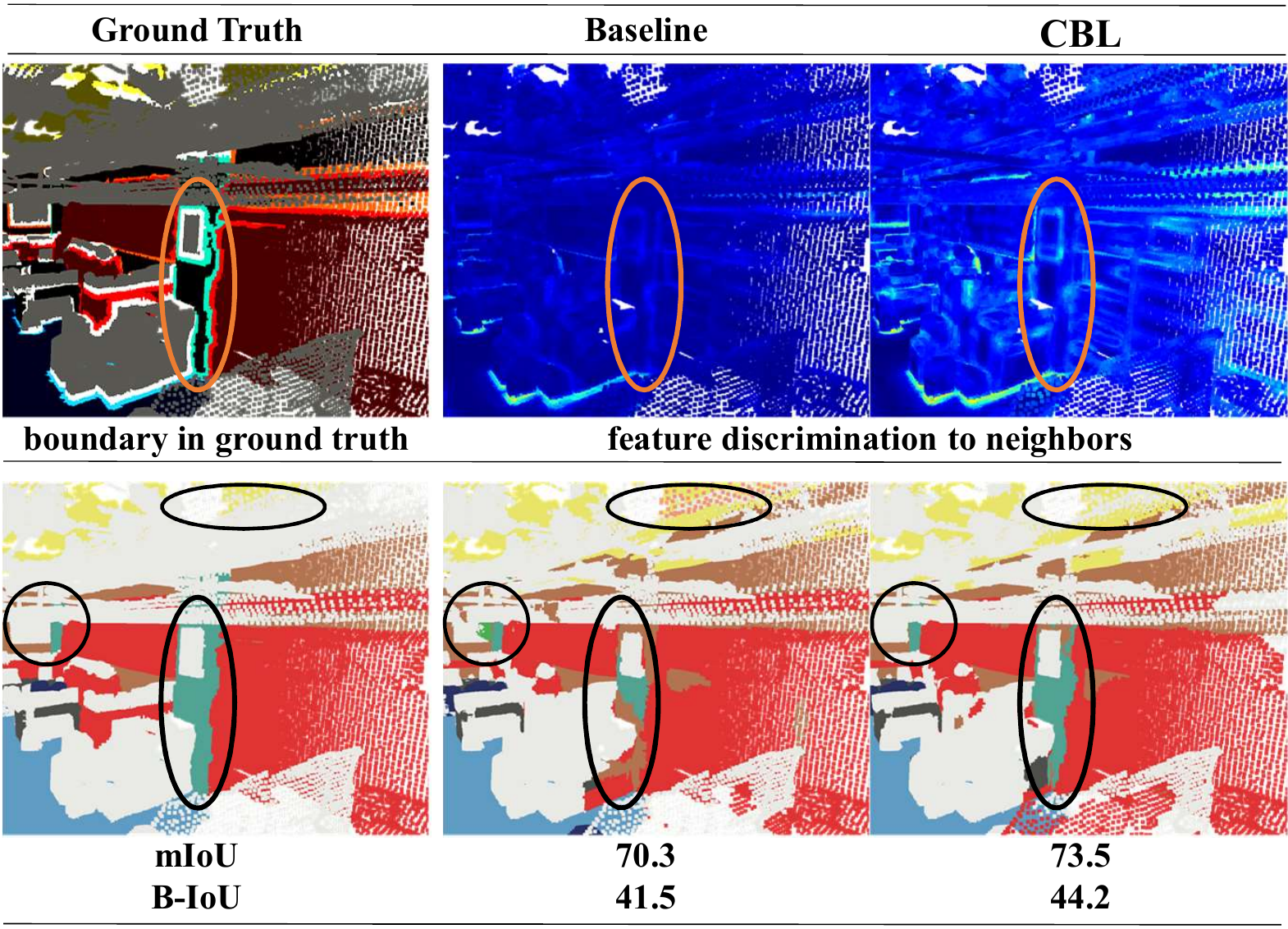}
\end{center}
   \caption{Contrastive Boundary Learning (top) discovers boundary from ground truth in each sub-sampled point cloud, \ie, sub-scene, through the sub-sampling procedure. By imposing contrastive optimization on boundary areas at multiple scales, CBL enhances the feature discrimination across boundaries (middle). Without an explicit boundary prediction, CBL improves boundary segmentation and achieves better scene segmentation results (bottom). The visualization is conducted on S3DIS testset Area 5.
   }
\label{fig:abstract}
\end{figure}

3D point cloud semantic segmentation aims to assign semantic categories to each 3D data point, while robust 3D segmentation is very important for various applications \cite{ptsurvey, ptSreview}, including autonomous driving, unmanned aerial vehicles, and augmented reality.

However, despite that various point cloud segmentation methods have been developed, little attention has been put on boundaries in 3D point clouds. Accurate segmentation on scene boundaries can be of great importance. Firstly, a clean boundary estimation can be beneficial for overall segmentation performance. For example, in 2D image segmentation, accurate segmentation on boundary is the key to generate high-fidelity masks \cite{bound_bio,bound_iou,bound_loss}. Secondly, compared to object categories that usually have a large portion of 3D points, 
such as buildings and trees, erroneous boundary segmentation could affect the recognition of object categories with much fewer points (\eg, pedestrians and pillars) to a greater extent. This can be particularly hazardous for applications like autonomous driving, \eg, crashing into curbs if boundaries are recognized inaccurately by a self-driving car.

Unfortunately, most previous 3D segmentation methods generally overlook the segmentation on scene boundaries. Though a few methods have considered boundaries, they still lack an explicit and comprehensive investigation to analyze the segmentation performance on boundary areas. They also perform unsatisfactorily on the overall segmentation performance.

Therefore, to deliver a more thorough study of the segmentation on boundaries, we first explore metrics to quantify the segmentation performance on scene boundaries. After revealing the unsatisfactory performance, we propose a novel Contrastive Boundary Learning (CBL) framework to help optimize the segmentation performance on boundaries particularly, which also consistently improves the overall performance for different baseline methods.

In particular, current popular segmentation metrics lack specific measurements on boundaries, making it difficult to reveal the boundary segmentation quality in existing methods. To make a clearer view on the performance on boundaries, we calculate the popular mean intersection-over-union (mIoU) for boundary areas and inner (non-boundary) areas separately. By comparing the performance on types of areas as well as the overall performance, the unsatisfactory performance on boundary areas can be directly revealed.
Moreover, to describe the performance on boundaries more comprehensively, we consider the alignment between the boundary in the ground truth and the boundary in model segmentation results. Therefore, we introduce the popular boundary IoU~\cite{bound_iou} score (B-IoU) used in 2D instance segmentation for evaluation, which also gives a much lower score compared with the overall performance in mIoU.

After identifying the boundary segmentation difficulties,
we further propose a novel contrastive boundary learning (CBL) framework to better align the boundaries of model predictions with ground-truth data's boundaries.
As shown in \cref{fig:abstract}, CBL optimizes a model on the feature representation of points in boundary areas, enhancing the feature discrimination across the scene boundaries. 
Furthermore, to make model better aware of the boundary areas at multiple semantic scales, we also develop a sub-scene boundary mining strategy, which leverages the sub-sampling procedure to discover boundary points in each sub-sampled point cloud, \ie, sub-scene.
Specifically, CBL operates across different sub-sampling stages and
facilitates 3D segmentation methods to learn better feature representation around boundary areas.

Empirically, we experiment with three baselines across four datasets. We first present the unsatisfactory performance on boundary areas when using current point cloud segmentation methods and then show that CBL can assist baseline in achieving promising boundary and overall performance. For example, the proposed CBL helps RandLA-Net surpass current state-of-the-art methods on the Semantic3D dataset and enables a basic ConvNet to achieve leading performance on the S3DIS dataset.

Our contributions are as follows:
\begin{itemize}
    \setlength\itemsep{-0.1em}
    \item We explore the boundary problem in current 3D point cloud segmentation and quantify it with metrics that consider boundary area, \eg, boundary IoU. The results reveal that current methods deliver much worse accuracy in boundary areas than their overall performance.

    \item We propose a novel Contrastive Boundary Learning (CBL) framework, which improves the feature representation by contrasting the point features across the scene boundaries. It thus improves the segmentation performance around boundary areas and subsequently the overall performance.

    \item We conduct extensive experiments and show that CBL can bring significant and consistent improvements on boundary area as well as overall performance across all baselines. These empirical results further demonstrate that CBL is effective for improving boundary segmentation performance, and accurate boundary segmentation is important for robust 3D segmentation.  
\end{itemize}

\begin{figure*}
\begin{center}
\resizebox{\linewidth}{!}{%
  \includegraphics[width=\linewidth]{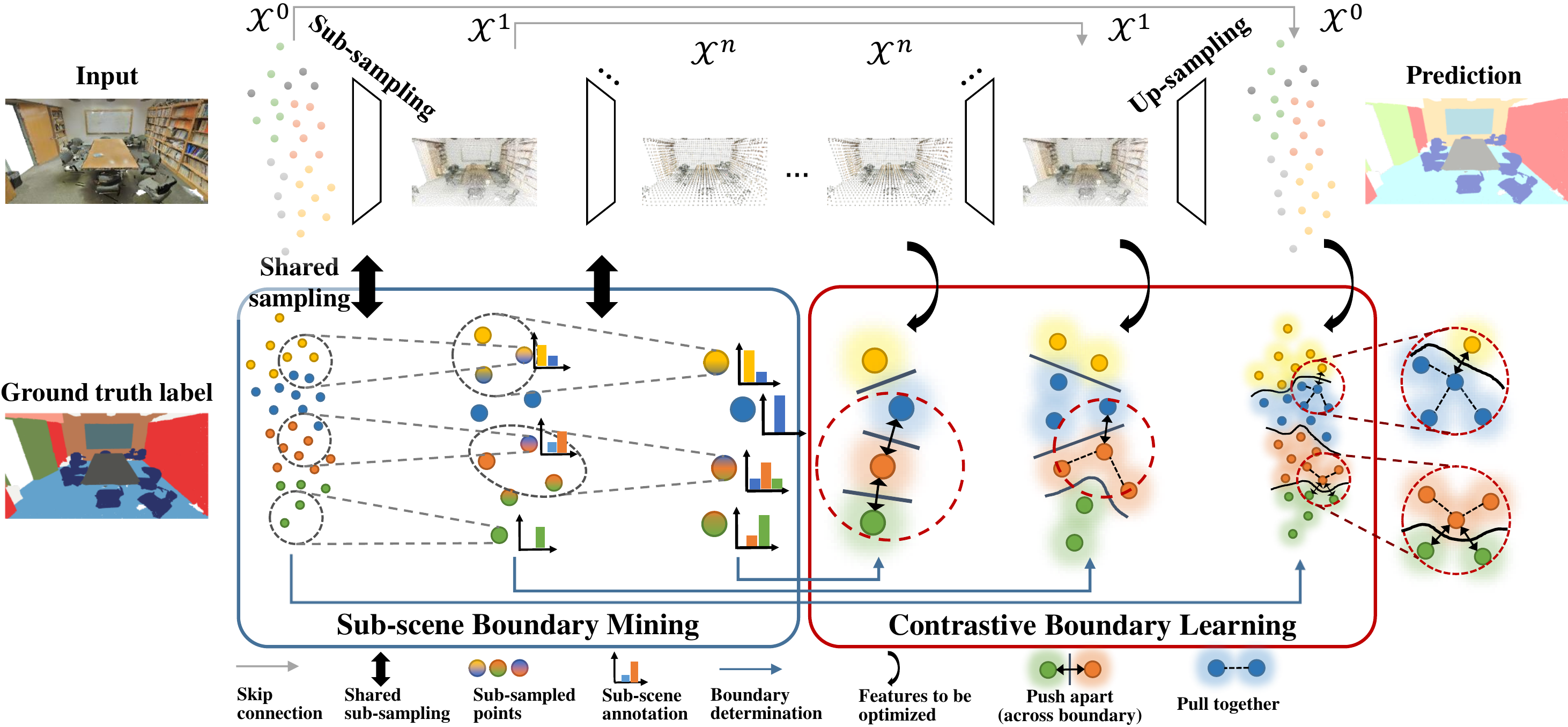}
}%
\end{center}
\caption{The detailed illustration of the Contrastive Boundary Learning.}
\label{fig:contrast}
\end{figure*}

\section{Related work}

\noindent\textbf{Point cloud segmentation.}
Point cloud semantic segmentation aims to assign semantic labels to each 3D point. Recent deep learning methods have taken over traditional methods~\cite{hf_hist, hf_struct} that use hand-crafted features, which can be roughly divided into projection-based and point-based methods.

Projection-based methods project 3D points to grid-like structure, either 2D image~\cite{seg_mv_SqueezeSegV3, seg_mv_RangeNet, vmvf, others_roaddet} or 3D voxels~\cite{Minkowski, ocnn, seg_vx_SEGCloud,occuseg}. 
For the 2D image plane, we can make use of existing studies for 2D image processing. However, a complete 3D segmentation generally requires taking multiple viewpoints and re-projection~\cite{seg_mv_snapNet, seg_mv_proj}, which may result in surface occlusions. For 3D voxels, sparse convolutions~\cite{seg_vx_SSCN, cls_vx_sparseConvSubmanifold,seg_vx_voxsegnet} are proposed to alleviate the resource consumption in voxel construction, considering the large emptiness in 3D space. In general, the voxel resolution incurs the trade-off between losing detail and being resource-demanding~\cite{cls_vox_APkernels}.
Point-based network directly operates on 3D points, while a pioneering work in this direction is PointNet~\cite{pointnet}, which uses point-wise MLPs to process per-point feature. Following this success, recent works adopt an encoder-decoder paradigm~\cite{pointnet++}. Various local aggregation modules are proposed to examine the local context in point clouds, including 3D convolution~\cite{fkaconv, pointcnn,pointconv,kpconv}, attentional operations~\cite{randlanet, pct, pttransformer}, and graph-based operation~\cite{dgcnn,spg}. To better process unstructured point cloud, sub-sampling~\cite{sample,PointASNL,pat,others_sasa}, up-sampling~\cite{seg_baaf,pyramidpoint}, and post-processing modules~\cite{bound_3d_cga,bound_3d_jse} are also considered to enhance point cloud representation.
Despite these developments in different modules, the boundary in point cloud segmentation has rarely been explored.

\noindent\textbf{Boundary in segmentation.}
Boundary problem has a long history in 2D image processing~\cite{bound_sample,bound_bio, bound_iou, bound_loss}, whereas only few works~\cite{bound_3d_pred,bound_3d_jse} 
realize the significance of boundary in 3D point cloud segmentation. However, both works involve complex modules for explicit boundary prediction~\cite{bound_3d_jse,bound_3d_pred} or local aggregation~\cite{bound_3d_jse}. These operations largely increase the model complexity, yet yield limited performance gain for overall metric. Regarding segmentation performance on boundaries, they also only give qualitative results.
In comparison, we present a contrastive learning framework that brings little overhead to the model and can improve upon various baselines with simple adaption.
Additionally, we would like to note that,
we for the first time,
quantify the boundary quality with numeric metric, and demonstrate that boundary problem is indeed widely existing across current methods.

\noindent\textbf{Contrastive learning.}
Contrastive learning~\cite{infonce, softnn, supcon, others_regioncl, moco, cont_simple} has shown promising performance in representation learning for computer vision tasks, ranging from unsupervised settings to supervised settings. In recent works, 
contrastive learning has also been introduced into 2D segmentation~\cite{imgcl_dense, imgcl_glb} as well as unsupervised representation learning in point cloud processing~\cite{pointcontrast, p4contrast, contrastscene}.
Especially, PointContrast~\cite{pointcontrast} conducts point-wise contrastive learning to overcome geometric transformation, such as rigid transformation. 
P4Contrast~\cite{p4contrast} suggests a more flexible contrasting strategy to promote multi-modal fusion between geometric and RGB information. 
In contrast, in our work, we take a supervised setting and demonstrate with CBL that contrastive learning is well-suited for improving segmentation quality on boundary areas. Additionally, unlike the above works that only use points at input point cloud, we utilize the sub-sampled point cloud to examine scene context at multiple scales.

\section{Segmentation on Boundaries}
\label{sec:metric}

Since most of the current works focus on the improvement of general metrics, such as mean intersection over union (mIoU), overall accuracy (OA), and mean average precision (mAP), the boundary quality in point cloud segmentation is usually overlooked. Unlike recent boundary-related works~\cite{bound_3d_pred,bound_3d_jse} that give only qualitative results on boundaries, we are the first to quantify the quality of segmentation on boundaries. Particularly, we introduce a series of metrics for presentation, including mIoU@boundary, mIoU@inner and the boundary IoU (B-IoU) score from 2D instance segmentation tasks~\cite{bound_iou}.

Based on ground-truths data, we consider a point as a boundary point if there exist points that have a different annotated label in its neighborhood. Similarly, for model predictions, we consider a point as a boundary point if there exist nearby points with a different predicted label.
More formally, we note the point cloud as $\mathcal X$ and the $i$-th point as $x_i$, whose local neighborhood is $\mathcal N_i = \mathcal N(x_i)$, corresponding ground truth label is $l_i$, and the model predicted label is $p_i$. We further note the set of boundary points in ground-truth as $\mathcal B_{l}$ and those in predicted segmentation as $\mathcal B_{p}$, thus we have:
\begin{equation}
\begin{array}{ll}
    \mathcal B_{l}  & \hspace{-8pt} = \{x_i \in \mathcal X \text{ } | \text{ } \exists ~ x_j\in \mathcal N_i, \text{ } l_j \neq l_i \}, \\
    \mathcal B_{p}  & \hspace{-8pt} = \{x_i \in \mathcal X \text{ } | \text{ } \exists ~ x_j\in \mathcal N_i, \text{ } p_j \neq p_i \},
\end{array}
\label{eq:bound}
\end{equation}
where we set $\mathcal N_i$ to be the radius neighborhood with a radius of $0.1$ following the common practice\cite{kpconv, closerlook}.

To examine the boundary segmentation results, an intuitive way is to calculate the mIoU within the boundary area, \ie, mIoU@boundary. To further compare the model performance in boundary and non-boundary (inner) area, we further calculate the mIoU in the inner area, \ie mIoU@inner.
Given that mIoU is calculated on the whole point cloud $\mathcal X$ as:
\begin{equation}
    \text{mIoU}(\mathcal X) = \frac 1K \sum_{k=1}^K \frac{\sum_{x_i\in\mathcal X} \mathbbm 1 [p_i = k \land l_i=k]} {\sum_{x_j\in\mathcal X} \mathbbm 1 [p_j=k \lor l_j=k]},
\label{eq:miou}
\end{equation}
where $K$ is the total number of classes and $\mathbbm{1}[\cdot]$ represents a boolean function that outputs 1 if the condition within $[\cdot]$ is true and 0 otherwise.
We have the mIoU@boundary and mIoU@inner defined as:
\begin{equation}
\begin{array}{ll}
\text{mIoU@boundary} &  \hspace{-8pt}= \text{mIoU}(\mathcal B_l), \\
\text{mIoU@inner} &  \hspace{-8pt}= \text{mIoU}(\mathcal X - \mathcal B_l),
\end{array}
\label{eq:bound_miou}
\end{equation}
where $\mathcal X - \mathcal B_l$ is the set of points in inner area.

However, the mIoU@boundary and mIoU@inner do not consider the false boundary in model predicted segmentation. Inspired by boundary IoU\cite{bound_iou} for 2D instance segmentation, for better evaluation, we consider the alignment between boundary in segmentation predictions and boundary in ground truth data. It thus leads to the following B-IoU for evaluation:
\begin{equation}
    \text{B-IoU} = \frac{|\mathcal B_l \cap \mathcal B_p|}{|\mathcal B_l \cup \mathcal B_p|}.
\label{eq:bound_iou}
\end{equation}

\section{Method}
In this section, we present our contrastive boundary learning (CBL) framework{\tly, shown in \cref{fig:contrast}}.
It imposes contrastive learning to enhance the feature discrimination across boundaries. Then, to deeply augment the model performance on boundaries, we enable the CBL in sub-sampled point clouds, \ie, sub-scene, through the sub-scene boundary mining.

\noindent\textbf{Contrastive Boundary Learning.}
We follow the widely used InfoNCE loss\cite{infonce} and its generalization~\cite{nce, softnn} to define the contrastive optimization goal on boundary points.
In particular, for a boundary point $x_i\in \mathcal B_l$, we encourage learned representations more similar to its neighbor points from the same category and more distinguished from other neighbor points from different categories, \ie,
\newcommand{\A}{\mathcal A}
\newcommand{\f}{f}  
\newcommand{\B}{\mathcal B}
\begin{equation}
    L_{CBL} = \frac {-1}{|B_l|} \sum_{x_i \in B_l} \log \frac
{ \displaystyle \sum_{ x_j\in \mathcal N_i \land l_j = l_i } \exp(-d(\f_i, \f_j) / \tau) }
{ \displaystyle \sum_{ x_k\in\mathcal N_i} \exp(-d(\f_i, \f_k) / \tau) },
\label{eq:cbl}
\end{equation}
where ${\f_i}$ is the feature of $x_i$, $d(\cdot, \cdot)$ is a distance measurement and $\tau$ is the temperature in contrastive learning.
The contrastive learning described by \cref{eq:cbl} focuses on boundary points only {\tly(the dashed circles in red in \cref{fig:contrast})}. First, we consider all the boundary points $\mathcal B_l$ from ground-truth data as defined in \cref{eq:bound}.
Then, for each point $x_i\in \B_l$, we restrict the sampling of its positive and negative points to be within its local neighborhood $\mathcal N_i$. With such strong spatial restriction, we obtain positive pairs for $x_i$ as $\{x_j\in \mathcal N_i \land l_j=l_i\}$, and other neighboring points, \ie $\{x_j\in \mathcal N_i \land l_j\neq l_i\}$, are negative pairs.
Therefore, the contrastive learning enhances the feature discrimination across scene boundaries, which is important for improving segmentation on boundary areas.

\noindent\textbf{Sub-scene Boundary Mining.}
To better explore scene boundaries,
we examine the boundaries in sub-sampled point clouds at multiple scales, which enables the contrastive boundary learning on different sub-sampling stages of a backbone model.
Collecting boundary points from the input point cloud is straightforward with the ground truth label. 
However, after sub-sampling, it is difficult to obtain a proper definition of boundary point set following \cref{eq:bound}, due to the undefined label for sub-sampled points\cite{omni}. Therefore, to enable CBL in sub-sampled point cloud, we propose the sub-scene boundary mining that determines the set of ground-truth boundary points in each sub-sampling stage. Specifically, we use superscripts to denote stage. At the sub-sampling stage $n$, we represent its sub-sampled point cloud as $\mathcal X^n$. For input point cloud, we have $\mathcal X^0 = \mathcal X$. When collecting a set of boundary points $\mathcal B^n_l \in \mathcal X^n$ in stage $n$, it is required to determine the label $l^n_i$ of a sub-sampled point $x^n_i\in\mathcal X^n$, \ie, the sub-scene annotation.
As each sub-sampled point $x_i^n\in \mathcal X^n$ is aggregated from a group of points in its previous point cloud $\mathcal X^{n-1}$; we thus utilize the sub-sampling procedure to determine the label iteratively. We take $l^0_i$ to be the one-hot label of ground truth label $l_i$ for point $x^0_i = x_i$, and have the following:
\begin{equation}
\begin{array}{ll}
    l^n_i &= \text{AVG}(\{ l^{n-1}_j | x^{n-1}_j \in \mathcal N^{n-1} (x^n_i) \}),
\end{array}
\label{eq:subscene:label}
\end{equation}
where $\mathcal N^{n-1} (x^n_i)$ denotes the local neighbors of $x^n_i$ in previous stage {\tly (the dashed circles in grey in \cref{fig:contrast})}, \ie, the group of points aggregated from $\mathcal X^{n-1}$ to be represented by the single point $x^n_i\in\mathcal X^n$ after sub-sampling procedure, and $\text{AVG}$ is the average-pooling.

With \cref{eq:subscene:label} and ground-truth labels, we can iteratively obtain the sub-scene annotation $l^n_i$ as a distribution, whose $k$-th location describes the proportion of $k$-th class in its corresponding group of points in the input point cloud. To determine the set of boundary points in sub-sampled point cloud $\mathcal X^n$, we simply take $\arg\max l^n_i$ to allow the evaluation of boundary point in \cref{eq:bound}\footnotemark\footnotetext{\tly We choose $\arg\max$ for its simplicity and non-parametric nature. We provide more analysis on this choice in the appendix.}, and use the feature of sub-sampled point for the contrastive boundary optimization in \cref{eq:cbl}. Finally, with sub-scene boundary mining, we have CBL applied at all stages and the final loss is
\begin{equation}
    L = L_\text{cross entropy} + \lambda \sum_{n} L^n_{CBL},
\end{equation} 
where $L^n_{CBL}$ is the CBL loss at stage $n$ and $\lambda$ is the loss weight.

\section{Implementation Details and Baselines}
\begin{figure}
\begin{center}
  \includegraphics[width=\linewidth]{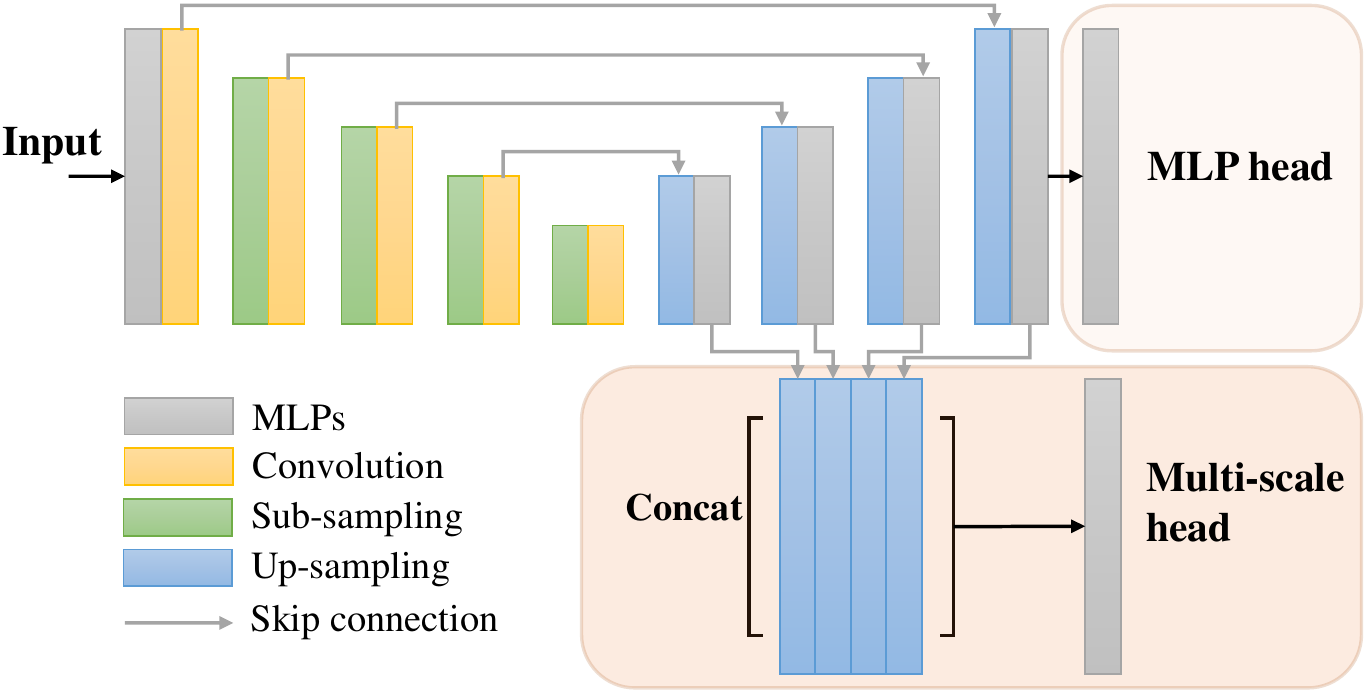}
\end{center}
   \caption{The architecture of the 3D ConvNet model, which follows the widely adopted encoder-decoder paradigm, with an optional multi-scale prediction head. More details are provided in the appendix.}
\label{fig:baseline}
\end{figure}
\label{sec:method:arch}
As 3D ConvNet has been a popular backbone model for point cloud processing, to present a generalized implementation, we illustrate with a ConvNet baseline (\cref{fig:baseline}) as a case study for applying CBL in point cloud processing.
Following \cite{convpoints, mtconv}, we build the ConvNet with convolution in 3D continuous space:
\begin{equation}
	\displaystyle
	f_i = ( h\circ g)(x_i) = \sum_{x_j\in\mathcal N_i} g(x_i-x_j) h(x_j),
	\label{eq:conv}
\end{equation}
where $\circ$ denotes convolution operator and the continuous kernel $g(\cdot)$ is approximated by one-layer MLP and set $h(x_j) = f_j$ to simply use the feature of point $x_j$. We note that the 3D convolution in \cref{eq:conv} is purely based on spatial location between the center point and its neighbors, compared to other advanced local aggregation modules that utilize the local context~\cite{randlanet, pttransformer}.

To better utilize the boundary features optimized by CBL at multiple scales, we use a multi-scale head for prediction, which simply concatenates the point feature from each sub-sampled point cloud into the last output layer. As we would show in the ablation study (\cref{sec:ablation}), such concatenation across multiple scales fails without the CBL. Note that CBL can be married to any other multi-stage backbone. Specifically, we also apply the CBL to two other popular baselines: the RandLA-Net\cite{randlanet} and CloserLook3D\cite{closerlook}, to demonstrate the generalizability. RandLA-Net leverages random sampling and attentive local aggregation to handle the large-scale scene with fast processing; CloserLook3D proposes a parameter-free PosPool module that largely reduces model parameters and resources consumption, while achieving comparable performance against other methods with parametric aggregation module, such as KPConv\cite{kpconv}. Together with the ConvNet baseline, our experiments cover the backbone with most of the typical local aggregation methods {\tly for} point cloud, ranging from convolution, attentional operation, to parameter-free operation.
{\tly For training, we follow the setup of baseline and set the loss weight $\lambda=0.1$. More details will be provided in the appendix.}

\begin{table}
\begin{center}
\centering
\resizebox{\linewidth}{!}{%
\begin{tabular}{r|ccc|c}
\hline
                                     & \multicolumn{3}{c|}{mIoU}                                                                   &                              \\
\multirow{-2}{*}{methods}            & overall                      & @boundary                    & @inner                       & \multirow{-2}{*}{B-IoU}      \\ \hline
pointnet\cite{pointnet}              & 41.1                         & 30.2                         & 53.4                         & 35.6                         \\
KPConv\cite{kpconv}                  & 67.3                         & 50.5                         & 71.1                         & 58.9                         \\
JSE-Net\cite{bound_3d_jse}*          & 67.7                         & 50.5                         & 71.4                         & 60.9                         \\
RandLA-Net\cite{randlanet}           & 62.6                         & 44.1                         & 65.8                         & 45.4                         \\
CloserLook3D\cite{closerlook}        & 66.9                         & 50.0                         & 70.7                         & 59.2                         \\
ConvNet                              & 67.4                         & 50.1                         & 71.2                         & 59.6                         \\ \hline
                                     & 65.3                         & 47.4                         & 67.2                         & 49.9                         \\
\multirow{-2}{*}{RandLA-Net + CBL}   & \cellcolor[HTML]{ECF4FF}+2.7 & \cellcolor[HTML]{ECF4FF}+3.3 & \cellcolor[HTML]{ECF4FF}+1.4 & \cellcolor[HTML]{ECF4FF}+4.5 \\
\hline
                                     & 67.5                         & 50.6                         & 71.0                         & 60.4                         \\
\multirow{-2}{*}{CloserLook3D + CBL} & \cellcolor[HTML]{ECF4FF}+0.6 & \cellcolor[HTML]{ECF4FF}+0.6 & \cellcolor[HTML]{ECF4FF}+0.3 & \cellcolor[HTML]{ECF4FF}+1.2 \\
\hline
                                     & 69.4                         & 52.6                         & 73.1                         & 61.5                         \\
\multirow{-2}{*}{ConvNet + CBL}      & \cellcolor[HTML]{ECF4FF}+2.0 & \cellcolor[HTML]{ECF4FF}+2.5 & \cellcolor[HTML]{ECF4FF}+1.9 & \cellcolor[HTML]{ECF4FF}+1.9                                 \\
\hline
\end{tabular}
}%
\end{center}
\caption{The results are obtained on the S3DIS datasets testset Area 5, following the instruction of the officially released code of each method. {\tly Method with * also consider boundaries.}
}
\label{tbl:bound}
\end{table}

\begin{table*}
\begin{center}
\resizebox{\linewidth}{!}{%
\begin{tabular}{r |c c c | c c c c c c c c c c c c c}
\hline
 methods & mIoU & OA & mACC & ceiling & floor & wall & beam & column & window & door & table & chair & sofa & bookcase & board & clutter \\
\hline
PointNet   \cite{pointnet}              & 41.1 & -    & 49.0 & 88.8 & 97.3 & 69.8 & 0.1 & 3.9  & 46.3 & 10.8 & 59.0 & 52.6 & 5.9  & 40.3 & 26.4 & 33.2  \\
SegCloud   \cite{seg_vx_SEGCloud}       & 48.9 & -    & 57.4 & 90.1 & 96.1 & 69.9 & 0.0 & 18.4 & 38.4 & 23.1 & 70.4 & 75.9 & 40.9 & 58.4 & 13.0 & 41.6  \\
PointCNN   \cite{pointcnn}              & 57.3 & 85.9 & 63.9 & 92.3 & 98.2 & 79.4 & 0.0 & 17.6 & 22.8 & 62.1 & 74.4 & 80.6 & 31.7 & 66.7 & 62.1 & 56.7  \\
SPGraph    \cite{spg}                   & 58.0 & 86.4 & 66.5 & 89.4 & 96.9 & 78.1 & 0.0 & \textbf{42.8} & 48.9 & 61.6 & 84.7 & 75.4 & 69.8 & 52.6 & 2.1  & 52.2  \\
PCT\cite{pct} & 61.3 & - & 67.7 & 92.5 & 98.4 & 80.6 & 0.0 & 19.4 & 61.6 & 48.0 & 76.6 & 85.2 & 46.2 & 67.7 & 67.9 & 52.3 \\
HPEIN \cite{pointedge}                  & 61.9 & 87.2 & 68.3 & 91.5 & 98.2 & 81.4 & 0.0 & 23.3 & \textbf{65.3} & 40.0 & 75.5 & 87.7 & 58.5 & 67.8 & 65.6 & 49.4  \\
MinkowskiNet \cite{Minkowski}           & 65.4 & -    & 71.7 & 91.8 & \textbf{98.7} & \textbf{86.2} & 0.0 & 34.1 & 48.9 & 62.4 & 81.6 & 89.8 & 47.2 & 74.9 & \textbf{74.4} & 58.6  \\
KPConv  \cite{kpconv}            & 67.1 & -    & 72.8 & 92.8 & 97.3 & 82.4 & 0.0 & 23.9 & 58.0 & 69.0 & 81.5 & \textbf{91.0} & 75.4 & 75.3 & 66.7 & 58.9  \\
JSENet\cite{bound_3d_jse}*        & 67.7 & - & - & 93.8 & 97.0 & 83.0 & 0.0 & 23.2 & 61.3 & 71.6 & 89.9 & 79.8 & 75.6 & 72.3 & 72.7 & 60.4 \\
CGA-Net\cite{bound_3d_cga}       & 68.6 & - & - & 94.5 & 98.3 & 83.0 & 0.0 & 25.3 & 59.6 & 71.0 & \textbf{92.2} & 82.6 & 76.4 & \textbf{77.7} & 69.5 & 61.5 \\
\hline
RandLA-Net     \cite{randlanet}         & 62.4 & 87.2 & 71.4 & 91.1 & 95.6 & 80.2 & 0.0 & 24.7 & 62.3 & 47.7 & 76.2 & 83.7 & 60.2 & 71.1 & 65.7 & 53.8 \\
\textbf{ + CBL}               & {\color{red}65.3} & {\color{red}87.5} & {\color{red}74.5} & {\color{red}92.2} & {\color{red}97.7} & {\color{red}81.0} & 0.0 & {\color{red}36.8} & 61.0 & 39.4 & {\color{red}78.1} & {\color{red}88.1} & \textbf{\color{red}81.4} & {\color{red}71.5} & {\color{red}68.7} & 52.6 \\
\hline
CloserLook3D\cite{closerlook} & 66.9 & 90.0 & 72.1 & 94.8 & 98.4 & 82.5 & 0.0 & 25.5 & 51.3 & 70.9 & 92.1 & 81.9 & 76.7 & 70.1 & 64.5 & 61.2 \\
\textbf{ + CBL} & {\color{red} 67.5} & {\color{red} 90.2} & {\color{red} 72.7} & \textbf{\color{red} 94.9} &{98.4} & {\color{red} 83.1} & 0.0 & {\color{red} 27.3} & {\color{red} 55.0} & {\color{red} 71.2} & 91.9 & {\color{red} 82.9} & 75.9 & {\color{red} 71.3} & 63.5 & 60.4 \\
\hline
ConvNet & 67.4 & 90.1 & 72.9 & 94.1 & 98.1 & 83.1 & 0.0 & 24.9 & 53.5 & \textbf{73.0} & 91.7 & 82.3 & 76.5 & 72.3 & 66.9 & 60.8 \\
\textbf{ + CBL} & \textbf{\color{red}69.4} & \textbf{\color{red}90.6} & \textbf{\color{red}75.2} & 93.9 & {\color{red}98.4} & {\color{red}84.2} & 0.0 & {\color{red}37.0} & {\color{red}57.7} & 71.9 & 91.7 & 81.8 & {\color{red}77.8} & {\color{red}75.6} & {\color{red}69.1} & \textbf{\color{red}62.9} \\
\hline
\end{tabular}
}%
\end{center}
\caption{
Quantitative results on S3DIS Area 5 dataset~\cite{s3dis}, showing the mean IoU (mIoU) overall accuracy (OA) and the mean accuracy (mACC).
The {\color{red}red} denotes improvement over baseline and the \textbf{bold} or \textbf{\color{red}bold} denotes the best performance. Method with * also consider boundaries in their design.
}
\label{tbl:s3dis}
\end{table*}

\begin{figure*}
\begin{center}
    \includegraphics[width=\linewidth]{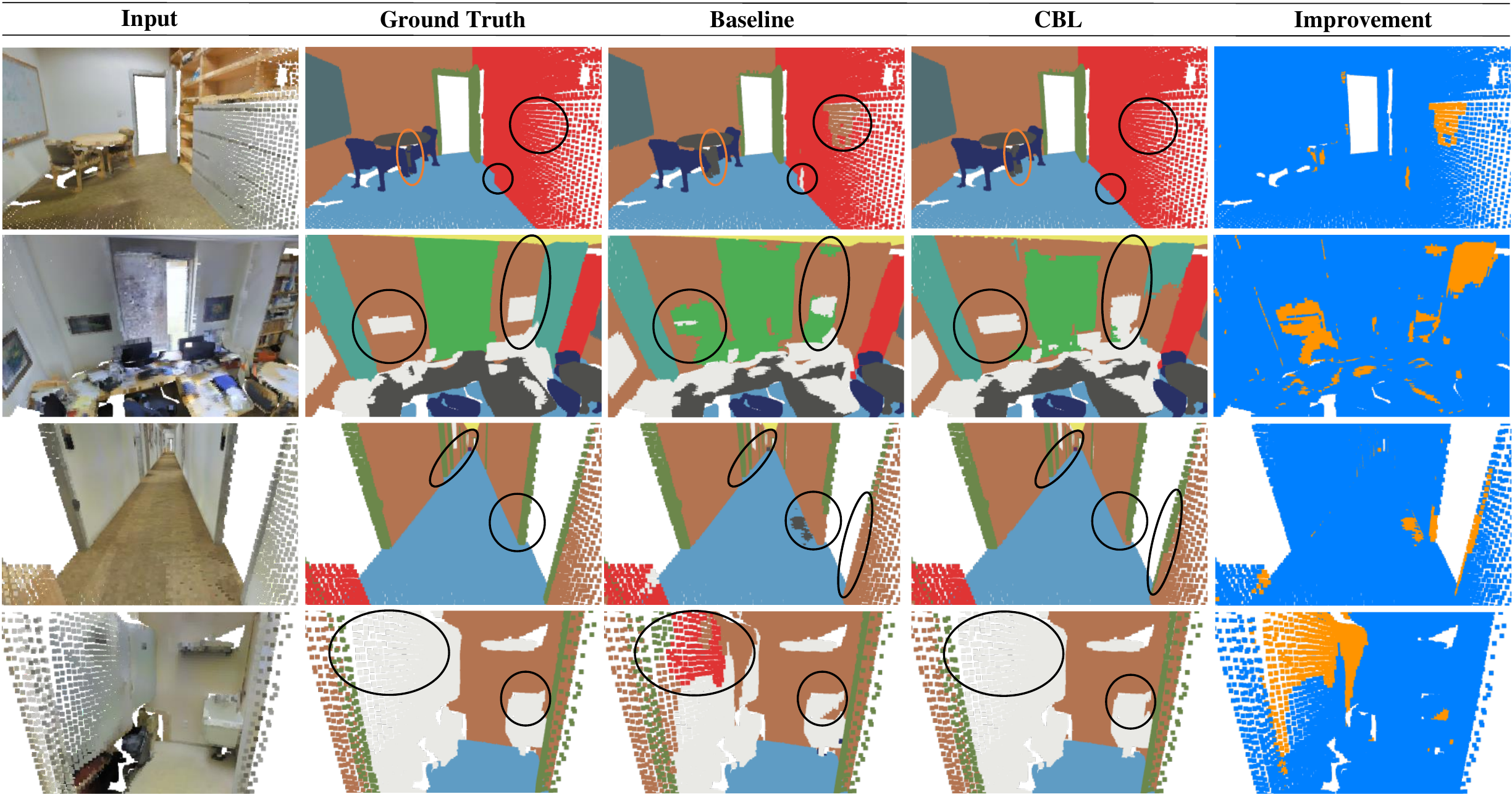}
\end{center}
   \caption{We compare the results of ConvNet baseline with CBL on several different scenes and show that the improvements are from boundaries. In offices (top 2), CBL can effectively improve the results on boundary areas, especially in a cluttered one (2nd row). In the last two rows (hallway and others), CBL avoids unnecessary boundaries, and repairs the missing boundary between walls and doors/objects at the right place.
   The visualization is done on S3DIS testset Area 5.
   }
\label{fig:demo}
\end{figure*}

\section{Experiments}

We first present the boundary problem with experiments. We then evaluate the benefits of the proposed CBL on multiple large-scale point cloud segmentation datasets, including in-door scenes (S3DIS~\cite{s3dis}, ScanNet~\cite{scannet}) and out-door scenes (Semantic3D~\cite{semantic3d}, NPM3D~\cite{npm3d}).

\subsection{The Boundary Problem in Experiment}
We experimentally compare the score given by mIoU, mIoU@boundary, mIoU@inner as well as the B-IoU. As shown in \cref{tbl:bound}, for recent 3D point cloud segmentation methods, the mIoU@boundary is much lower than the mIoU@inner. With the overall performance sitting between these two scores, it suggests that it is the boundary area that degenerates the overall segmentation performance. Similarly, B-IoU also agrees with the mIoU@boundary by giving a score that is far lagged behind the general performance of mIoU score. Hence, such observation indicates the unsatisfied segmentation quality on boundary areas.
While with the proposed CBL, the improvement on both mIoU@boundary and B-IoU is larger than the improvement on overall mIoU as well as the mIoU@inner, across all three baselines.
Due to the limited space, we provide more thorough studies in presenting the boundary problem in the appendix. 

\begin{table*}
\begin{center}
\resizebox{\linewidth}{!}{%
\begin{tabular}{r |c c c | c c c c c c c c c c c c c}
\hline
 methods & mIoU & OA & mACC & ceiling & floor & wall & beam & column & window & door & table & chair & sofa & bookcase & board & clutter \\
\hline
PointNet\cite{pointnet}           & 47.6 & 78.6 & 66.2 & 88.0 & 88.7 & 69.3 & 42.4 & 23.1 & 47.5 & 51.6 & 54.1 & 42.0 & 9.6 & 38.2 & 29.4 & 35.2 \\
RSNet\cite{seg_vx_RSNet}          & 56.5 & -    & 66.5 & 92.5 & 92.8 & 78.6 & 32.8 & 34.4 & 51.6 & 68.1 & 59.7 & 60.1 & 16.4 & 50.2 & 44.9 & 52.0 \\
SPG \cite{spg}                    & 62.1 & 86.4 & 73.0 & 89.9 & 95.1 & 76.4 & 62.8 & 47.1 & 55.3 & 68.4 & 73.5 & 69.2 & 63.2 & 45.9 & 8.7 & 52.9 \\
PointCNN\cite{pointcnn}           & 65.4 & 88.1 & 75.6 & \textbf{94.8} & \textbf{97.3} & 75.8 & 63.3 & 51.7 & 58.4 & 57.2 & 71.6 & 69.1 & 39.1 & 61.2 & 52.2 & 58.6 \\
PointWeb\cite{cls_ops_PointWeb}   & 66.7 & 87.3 & 76.2 & 93.5 & 94.2 & 80.8 & 52.4 & 41.3 & 64.9 & 68.1 & 71.4 & 67.1 & 50.3 & 62.7 & 62.2 & 58.5 \\
ShellNet\cite{cls_ops_ShellNet}   & 66.8 & 87.1 & -    & 90.2 & 93.6 & 79.9 & 60.4 & 44.1 & 64.9 & 52.9 & 71.6 & \textbf{84.7} & 53.8 & 64.6 & 48.6 & 59.4 \\
RandLA-Net\cite{randlanet}        & 70.0 & 88.0 & 82.0 & 93.1 & 96.1 & 80.6 & 62.4 & 48.0 & 64.4 & 69.4 & 69.4 & 76.4 & 60.0 & 64.2 & 65.9 & 60.1 \\
KPConv\cite{kpconv}               & 70.6 & -    & 79.1 & 93.6 & 92.4 & 83.1 & 63.9 & 54.3 & 66.1 & 76.6 & 57.8 & 64.0 & 69.3 & \textbf{74.9} & 61.3 & 60.3 \\
SCF-Net\cite{seg_ops_scf}         & 71.6 & 88.4 & 82.7 & 93.3 & 96.4 & 80.9 & \textbf{64.9} & 47.4 & 64.5 & 70.1 & 71.4 & 81.6 & 67.2 & 64.4 & \textbf{67.5} & 60.9 \\
BAAF\cite{seg_baaf}               & 72.2 & 88.9 & \textbf{83.1} & 93.3 & 96.8 & 81.6 & 61.9 & 49.5 & 65.4 & 73.3 & 72.0 & 83.7 & 67.5 & 64.3 & 67.0 & \textbf{62.4} \\
\hline
ConvNet                          & 69.7 & 88.6 & 76.8 & 93.8 & 91.9 & 84.2 & 46.3 & 52.1 & 66.7 & \textbf{78.5} & 75.2 & 72.8 & 70.1 & 71.7 & 57.1 & 61.3 \\
\textbf{ + CBL}           & 
\textbf{\color{red}73.1} & \textbf{\color{red}89.6} & {\color{red}79.4} & {\color{red}94.1} & {\color{red}94.2} & \textbf{\color{red}85.5} & {\color{red}50.4} & \textbf{\color{red}58.8} & \textbf{\color{red}70.3} & 78.3 & \textbf{\color{red}75.7} & {\color{red}75.0} & \textbf{\color{red}71.8} & {\color{red}74.0} & {\color{red}60.0} & \textbf{\color{red}62.4}

 \\ 
\hline

\end{tabular}
}%
\end{center}
\caption{
Quantitative results on S3DIS~\cite{s3dis} with 6-fold cross validation. The {\color{red}red} denotes improvement over baseline and the \textbf{bold} or \textbf{\color{red}bold} denotes the best performance.
}
\label{tbl:s3dis_cross}
\end{table*}

\begin{table*}
\begin{center}
\resizebox{\linewidth}{!}{%
\begin{tabular}{r |cc|ccccccccccccccccc }
\hline
 methods & mIoU (\%) & OA (\%) & man-made. & natural. & high veg. & low veg. & buildings & hard scape & scanning art. & cars \\
\hline
SnapNet       \cite{seg_mv_snapNet}   & 59.1  & 88.6   & 82.0   & 77.3   & 79.7   & 22.9   & 91.1   & 18.4   & 37.3   & 64.4  \\
SEGCloud      \cite{seg_vx_SEGCloud}   & 61.3  & 88.1   & 83.9   & 66.0   & 86.0   & 40.5   & 91.1   & 30.9   & 27.5   & 64.3  \\
SPG           \cite{spg}   & 73.2  & 94.0   & 97.4   & 92.6   & 87.9   & 44.0   & 83.2   & 31.0   & 63.5   & 76.2  \\
RGNet         \cite{rgnet} & 74.7	& 94.5    & \textbf{97.5}	& 93.0	& \textbf{88.1}	& 48.1	& 94.6	& 36.2	& 72.0	& 68.0 \\
KPConv        \cite{kpconv}   & 74.6  & 92.9   & 90.9   & 82.2   & 84.2   & 47.9   & 94.9   & 40.0   & \textbf{77.3}   & \textbf{79.7}  \\
RFCR\cite{omni}	            &  77.8   & 94.3   & 94.2   & 89.1   & 85.7   & 54.4   & 95.0	 &43.8    & 76.2 & 83.7 \\
SCF-Net\cite{seg_ops_scf}   &  77.6 &   94.7 &   97.1 &   \textbf{91.8} &   86.3 &   51.2 &   95.3 &   50.5 &   67.9 &   80.7 \\
\hline
ConvNet         &   72.8            &   92.6            &	92.2            &   79.9            &   84.4    &	41.3            &	95.2    &	41.2    &	62.6            &   85.6 \\
\textbf{ + CBL} &   {\color{red}75.0}	&   {\color{red}94.0}	&	{\color{red}96.2}	&   {\color{red}90.1}	&   84.0	&   {\color{red}47.5}	&   94.7	&   36.0	&   {\color{red}64.8}	&   {\color{red}86.3} \\
\hline
RandLA-Net    \cite{randlanet}   & 77.4  & 94.8   & 95.6   & 91.4   & 86.6   & 51.5   & 95.7   & 51.5   & 69.8   & 76.8 \\
\textbf{ + CBL} & \textbf{\color{red} 78.4} & \textbf{\color{red} 95.0} & 95.3   & 91.3   & {\color{red} 87.9}   & \textbf{\color{red} 55.6}   & \textbf{\color{red} 96.3}   & \textbf{\color{red} 56.2}   & 65.9   & {\color{red}78.2} \\
\hline
\end{tabular}
}%
\end{center}
\caption{Quantitative results on Semantic3D reduced-8 benchmark~\cite{semantic3d}. The metrics shown the mean IoU (mIoU) and overall accuracy (OA) obtained from benchmark site with only the recent published works included. The {\color{red}red} denotes improvement over baseline and the \textbf{bold} or \textbf{\color{red}bold} denotes the best performance.}
\label{tbl:semantic3d}
\end{table*}

\subsection{Performance Comparison}
\noindent\textbf{S3DIS Indoor Scene Segmentation.}
S3DIS~\cite{s3dis} is a challenging point cloud dataset of indoor scenes. It contains 3D RGB point clouds of 6 indoor areas covering 272 rooms. Each point is annotated with one of the 13 semantic categories, e.g., ceiling, floor, clutter.
As shown in \cref{tbl:s3dis}, our methods consistently improve across all three baselines, showing to be effective with different local aggregation modules.
Notably, the improvements are much more significant in classes, such as column (+13 compared to ConvNet baseline), than in other classes with large areas, such as wall and ceiling. Such observation shows our effectiveness on boundary areas; and with the consistent improvement across different classes, it also suggests that the CBL is NOT trading off between scenes of major and minor classes, but is indeed separating them more clearly. With the benefit of a cleaner boundary, the ConvNet finally achieves a leading performance of $69.4$ in mIoU.

We further demonstrate qualitatively in \cref{fig:demo} that, the CBL effectively improves the overall performance by improving segmentation on boundary areas.
Compared with JSENet\cite{bound_3d_jse} that also considers boundaries, we demonstrate our superiority by obtaining a much larger relative improvement to our baselines than that made by JSENet on its baseline, \ie, KPConv\cite{kpconv}, especially in classes that boundaries are important, \eg, column, window, sofa, bookcase and clutter, as well as the overall performance.
To avoid overfitting on S3DIS Area 5, we further conduct the 6-fold cross-validation, with the result reported in \cref{tbl:s3dis_cross}. A large improvement is also shown in column (+9.5), and consistent improvement is made across all classes except one (-0.2). Therefore, the proposed CBL can be indeed regarded as a general and effective method, achieving $73.1$ in mIoU with a common ConvNet baseline.

\noindent\textbf{Semantic3D Outdoor Scene Segmentation.}
In addition to improvement on S3DIS\cite{s3dis}, we demonstrate the generalizability across different types of scenes by evaluating CBL on point cloud collected at the outdoor environment, the Semantic3D~\cite{semantic3d} dataset. It is a large-scale dataset comprising over 4 billion points and provides 15 large point clouds for training, with each point annotated to one of the 8 classes, \eg, cars, buildings.
We use the reduced-8 benchmark and present the quantitative results in \cref{tbl:semantic3d}. We evaluate with both ConvNet and RandLA-Net\cite{randlanet} as baselines and observe consistent improvements. Especially, RandLA-Net has achieved state-of-the-art performance on multiple outdoor datasets and the improvement made on it can better demonstrate the effectiveness of our CBL. Notably, significant improvement is made in the high vegetation and low vegetation class, which are two classes that confuse most of the other methods. It is because the high/low vegetation usually co-exists at a near spatial distance and has a similar appearance, \eg, trees surrounded by bushes/grass, which makes the separation of these two scenes challenging. The large improvement in both of these two classes demonstrates the effective improvement on scene boundaries. Lastly, with CBL, RandLA-Net obtains a leading performance of $78.4$ in mIoU.

\begin{table}
\begin{center}
\resizebox{0.8\linewidth}{!}{%
\begin{tabular}{r|c|c}
\hline
methods                         & modality                    & mIoU (\%) \\ \hline
DCM-Net\cite{dcmnet}            & \multirow{2}{*}{3D + Mesh}  & 65.8      \\
VMNet\cite{vmnet}               &                             & 74.6      \\ \hline
SparseConvNet\cite{seg_vx_SSCN} & \multirow{5}{*}{3D (voxel)} & 72.5      \\
MinkowskiNet\cite{Minkowski}    &                             & 73.6      \\
O-CNN\cite{ocnn}                &                             & 76.2      \\
OccuSeg\cite{occuseg}           &                             & 76.4      \\
Mix3D\cite{mix3d}               &                             & 78.1      \\ \hline
BA-GEM\cite{bound_3d_pred}* & \multirow{7}{*}{3D (point)} & 63.5 \\
PointConv\cite{pointconv}       &                             & 66.6      \\
PointASNL\cite{PointASNL}       &                             & 66.6      \\
KP-Conv\cite{kpconv}            &                             & 68.4      \\
FusionNet\cite{fusionnet}       &                             & 68.8      \\
JSENet\cite{bound_3d_jse}*       &                             & 69.9      \\
RFCR\cite{omni}                 &                             & 70.2      \\ \hline
ConvNet                         & \multirow{2}{*}{3D (point)} & 69.1      \\
\textbf{ + CBL}          &                             & 70.5      \\
\hline
\end{tabular}
}%
\end{center}
\caption{Quantitative results on ScanNet~\cite{scannet} benchmark. Performance is taken from the official benchmark site by the time of submission. Methods with * also consider boundaries.
}
\label{tbl:scannet}
\end{table}

\begin{table}
\begin{center}

\begin{tabular}{r |c}
\hline
  methods  & mIoU (\%) \\
\hline
HDGCN\cite{hdgcn}	            & 68.3 \\
ConvPoint\cite{convpoints}	    & 75.9 \\
RandLANet\cite{randlanet}	    & 78.5 \\
KP-Conv\cite{kpconv}	        & 82.0 \\
FKAConv\cite{fkaconv}	        & 82.7 \\
PyramidPoint\cite{pyramidpoint} & 82.9 \\
\hline
ConvNet                 & 76.2 \\
\textbf{ + CBL}  & 78.6 \\
\hline
\end{tabular}

\end{center}
\caption{Quantitative results on Paris-Lille-3D of NPM3D~\cite{npm3d} benchmark, results obtained from online benchmark site by the time of submission.
}
\label{tbl:npm3d}
\end{table}

\noindent\textbf{Further experiments on NPM3D and ScanNet.}
To further demonstrate the generalization of the proposed CBL, we report on another two popular dataset, the ScanNet\cite{scannet} (indoor scene) and NPM3D\cite{npm3d} (outdoor scene). As shown in \cref{tbl:scannet} and \cref{tbl:npm3d}, our method achieves reasonable results and consistent improvement over the baseline. It thus shows that CBL is robust to different baselines, datasets, and types of scenes. Detailed results are available in the appendix.

\begin{table}
\begin{center}
\resizebox{\linewidth}{!}{%
\begin{tabular}{c|cc|l|l}
\hline
\multicolumn{1}{l|}{}                                                                      & \multicolumn{2}{c|}{CBL}                                                & \multicolumn{1}{l|}{}                                &                                                       \\
\multicolumn{1}{l|}{}                                                                      & @input                        & @sub-scenes                        & \multicolumn{1}{l|}{\multirow{-2}{*}{mIoU(\%)}}      & \multirow{-2}{*}{OA(\%)}                              \\ \hline
                                                                                           &                                    &                                    & 69.71 \hspace{10pt} -                                 & 88.97 \hspace{10pt} -                                  \\
                                                                                           & \cellcolor[HTML]{ECF4FF}\checkmark & \cellcolor[HTML]{ECF4FF}           & \cellcolor[HTML]{ECF4FF}70.05   \hspace{2pt}  {\color{red}+0.34}  & \cellcolor[HTML]{ECF4FF}89.01   \hspace{2pt}  {\color{red}+0.04}   \\
\multirow{-3}{*}{ConvNet}                                                                  & \cellcolor[HTML]{ECF4FF}\checkmark & \cellcolor[HTML]{ECF4FF}\checkmark & \cellcolor[HTML]{ECF4FF}70.98   \hspace{2pt}  {\color{red}+1.27}  & \cellcolor[HTML]{ECF4FF}89.31   \hspace{2pt}  {\color{red}+0.34}   \\ \hline
                                                                                           &                                    &                                    & 69.83   \hspace{2pt}  {\color{red}+0.12}                          & 88.88   \hspace{2pt}  -0.09                                        \\
\multirow{-2}{*}{\begin{tabular}[c]{@{}c@{}}ConvNet\\ (multiscale head)\end{tabular}} & \cellcolor[HTML]{DAE8FC}\checkmark & \cellcolor[HTML]{DAE8FC}\checkmark & \cellcolor[HTML]{DAE8FC}71.33   \hspace{2pt}  {\color{red}+1.62}  & \cellcolor[HTML]{DAE8FC}89.40   \hspace{2pt}  {\color{red}+0.43}   \\ \hline
\end{tabular}
}%
\end{center}
\caption{Results on validation set of ScanNet\cite{scannet}. The CBL @input refers to only conduct contrastive boundary learning on the input point cloud (with point feature extracted from last upsampled stage), and @sub-scene refers to the CBL with sub-scene boundary mining. The {\color{red}red} indicates relative improvement.}
\label{tbl:ablation}
\end{table}

\subsection{Ablation Studies}
\label{sec:ablation}
We conduct ablation studies on the ScanNet validation set to evaluate the effectiveness of different components in the proposed CBL scheme.

\noindent{\textbf{The Effectiveness of CBL.}} As shown in \cref{tbl:ablation}, the direct application of CBL on the input point cloud (without sub-scene boundary mining) can improve the performance, which demonstrates that boundary areas are worth more attention. By introducing sub-scene boundary mining, a more significant improvement is gained, as boundaries at multiple scales are identified and optimized in the CBL.

\noindent{\textbf{The Effect of Multi-scale Head.}}
Comparing the ConvNet baseline with and without the multi-scale head, we find that a direct application of multi-scale head can even hurt the performance (-0.09 in OA). It shows that a direct concatenation across multiple scales can not bring much benefit. In contrast, with multi-scale head, ConvNet with CBL is further boosted to gain a larger improvement in both mIoU and OA. It shows that the main improvement is originated from the more discriminative features learned by CBL at different sub-sampled point clouds.

\section{Conclusion}
In this paper, we comprehensively analyze the segmentation performance on scene boundaries for the current point cloud segmentation methods. We show that the current segmentation accuracy on boundaries is unsatisfactory and quantitatively present the boundary problem with metrics, including mIoU@boundary and B-IoU. We further propose Contrastive Boundary Learning (CBL) to explicitly optimize the feature on boundaries and improve the model performance on boundaries. The leading performance and consistent improvement across various baselines and datasets demonstrate the effectiveness of CBL and the importance of scene boundaries in 3D point cloud segmentation.

\noindent\textbf{Limitation and future work.}
One of our limitation is that we mainly concentrate on the scene boundaries while ignoring the broad inner areas. Therefore, in the future, we would like to further explore the role of boundary in point cloud segmentation and its relation with inner areas.

\noindent\textbf{Acknowledgement.}
Dr Baosheng Yu and Mr Liyao Tang are supported by ARC FL-170100117, and Dr Zhe Chen is supported by ARC IH-180100002.


{\small
\bibliographystyle{ieee_fullname}
\bibliography{arxiv}
}

\onecolumn
\newpage
\twocolumn

\appendix

\begin{figure*}
\centering
    \includegraphics[width=\linewidth]{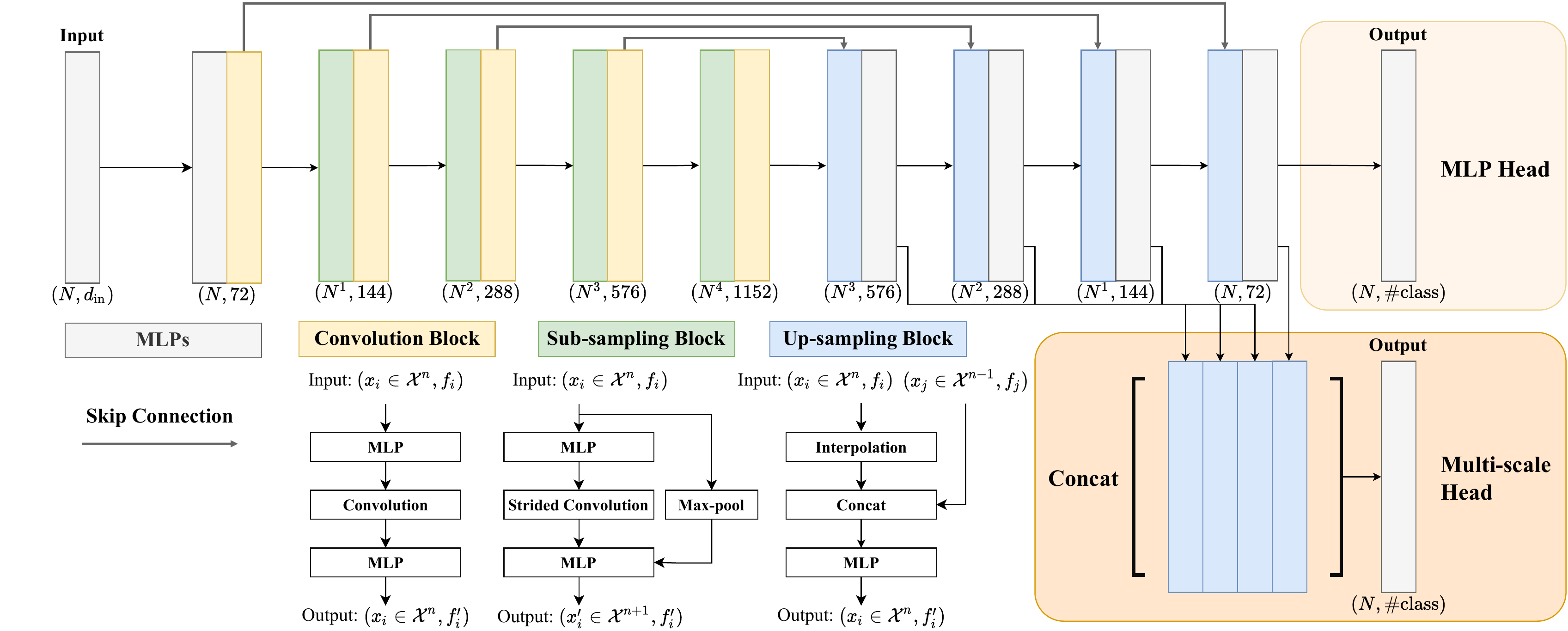}
    \caption{The detail architecture of ConvNet baseline.}
    \label{fig:baseline_fine}
\end{figure*}

\section{Introduction.}
In this supplementary material, we provide more details regarding baseline architecture (\cref{sec:sup:baseline}), the boundary problem \cref{sec:sup:bound}, visualization results (\cref{sec:sup:vis}), the training setup (\cref{sec:sup:train}), the effect of temperature (\cref{sec:sup:ablation:temperature}), the effect of design regarding sub-scene annotation (\cref{sec:sup:argmax}), and experiment results (\cref{sec:sup:rst}).

Especially, CBL achieves a new stat-of-the-art on S3DIS with the newly released transformer model (\cref{tbl:s3dis:pttrans}). 

\section{Architecture of ConvNet Baseline}
\label{sec:sup:baseline}
We show the specific architecture of our ConvNet baseline in \cref{fig:baseline_fine}. With a consistent notation, $\mathcal X^n$ is the point cloud in sub-sampling stage $n$, $f_i$ is the feature of point $x_i$, and $N^n = |\mathcal X^n|$ with $N=N^0$. We use the multi-scale head on all baselines when adapting the CBL.

\section{Further Analysis on Boundary Problem}
\label{sec:sup:bound}
We further account for the type of areas and class-specific analysis for better exploring the boundary problem. Specifically, we provide per-class IoU score that is separately calculated on boundary area $\mathcal B_l$ and inner area $\mathcal X-\mathcal B_l$.

As shown in \cref{tbl:bound_inner_iou}, we evaluate for all three baselines with and without the proposed CBL.
We notice that, large improvements are made on small objects, \eg column, which aligns with the observation  in \cref{tbl:s3dis} in main paper. We would like to add that, despite that CBL focuses only on boundaries, improvements are also made on inner area. We hypothesize the reason might be that the false boundary in model predicted segmentation is restrained, as features in inner area implicitly becomes more similar when the features across boundaries are optimized to be more distinctive by the CBL.

Moreover, for all three baselines, the improvement on boundary area is much more than that made on inner area, which is summarized in \cref{tbl:improve_bound_inner}.

Therefore, with metrics separately calculated on boundary and inner area, we clearly see that the improvement brought by CBL is mainly from the boundary areas. Such observation further emphasizes the importance of clear scene boundaries in point cloud segmentation task.

\section{More Visualizations}
\label{sec:sup:vis}
\begin{figure*}
\centering
  \begin{subfigure}{\linewidth}
    \includegraphics[width=\linewidth]{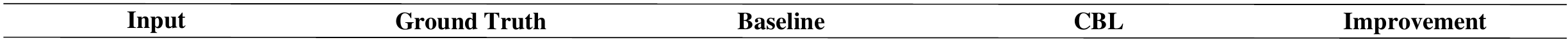}
    \includegraphics[width=\linewidth]{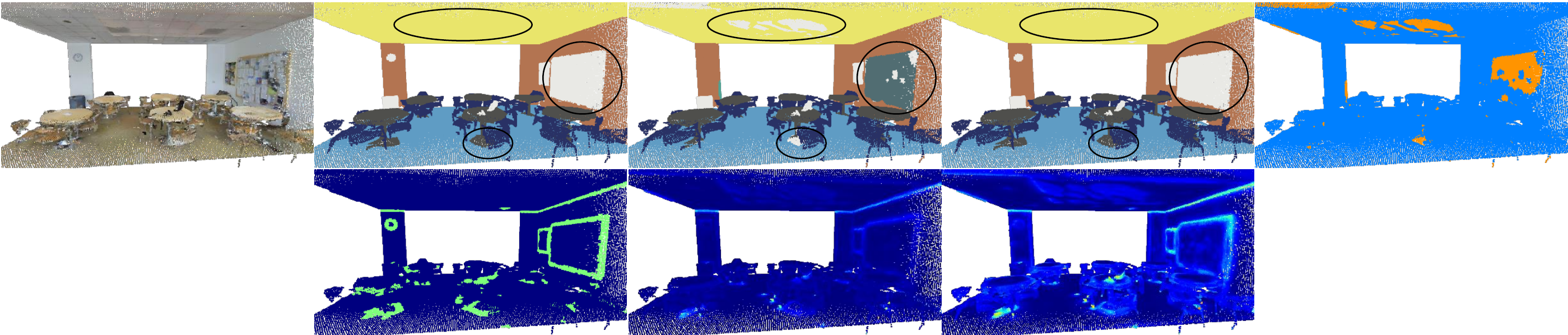}
  \caption{}
  \end{subfigure}
  \begin{subfigure}{\linewidth}
    \includegraphics[width=\linewidth]{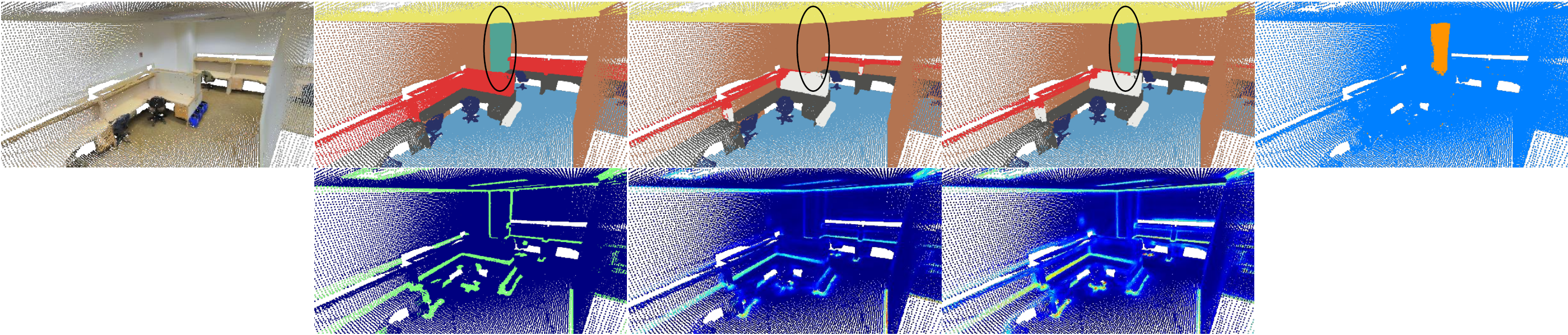}
  \caption{}
  \end{subfigure}
\caption{Large rooms.
We compare the results of ConvNet baseline with CBL. On the every second row, we visualize the boundary points calculated from the ground truth label, and the feature discrimination among neighboring points for each model.
The improvement on the first row and the enhanced feature discrimination on the second row show that CBL improves the features across boundaries to obtain a better segmentation quality on boundary areas.
The visualization is done on S3DIS testset Area 5.
}
\label{fig:demo_more_room}
\end{figure*}

\begin{figure*}  
\centering
  \begin{subfigure}{\linewidth}
    \includegraphics[width=\linewidth]{plot_more/demo_title.pdf}
    \includegraphics[width=\linewidth]{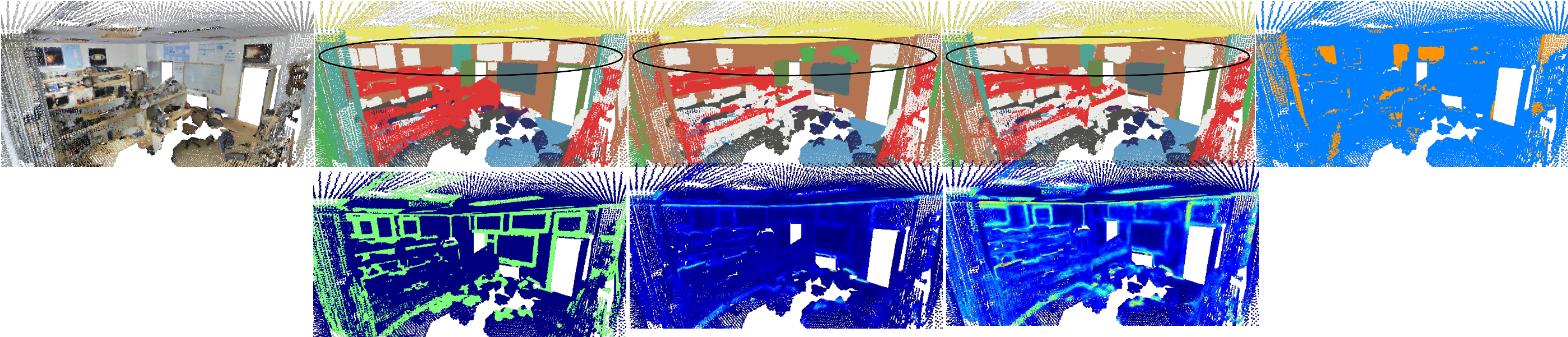}
  \end{subfigure}
  \begin{subfigure}{\linewidth}
    \includegraphics[width=\linewidth]{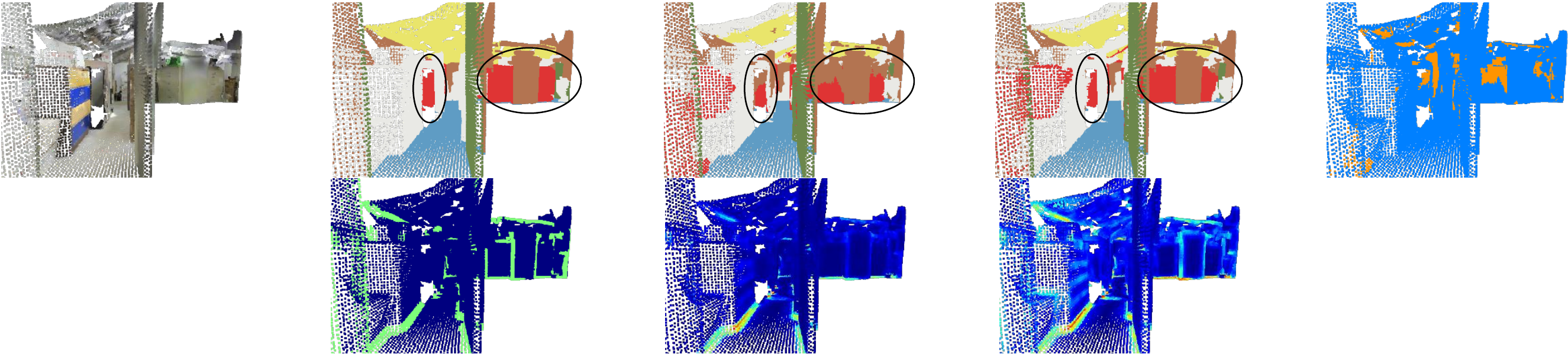}
  \end{subfigure}
  \begin{subfigure}{\linewidth}
    \includegraphics[width=\linewidth]{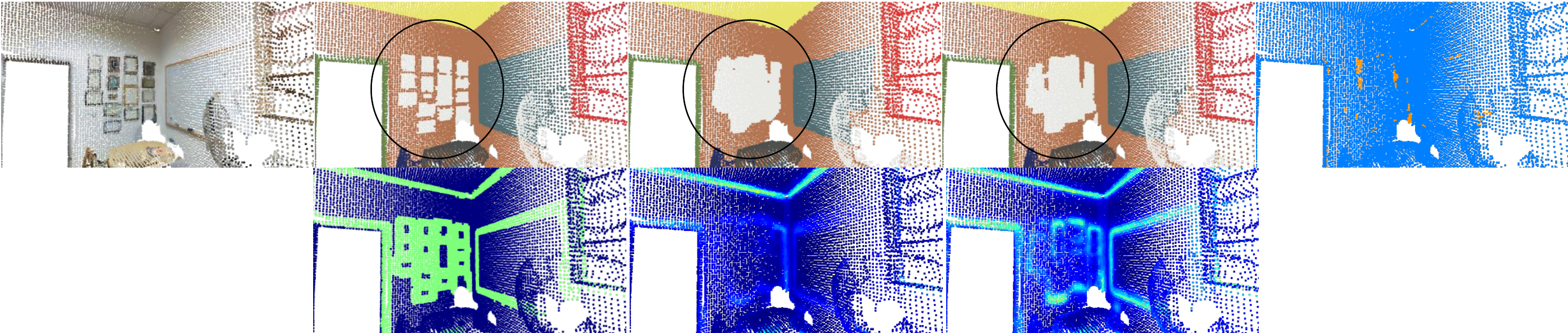}
  \end{subfigure}

\caption{Cluttered space. Same as above (\cref{fig:demo_more_room}).}
\label{fig:demo_more_clutter}

\end{figure*}

\begin{figure*}
\centering
  \begin{subfigure}{\linewidth}
    \includegraphics[width=\linewidth]{plot_more/demo_title.pdf}
    \includegraphics[width=\linewidth]{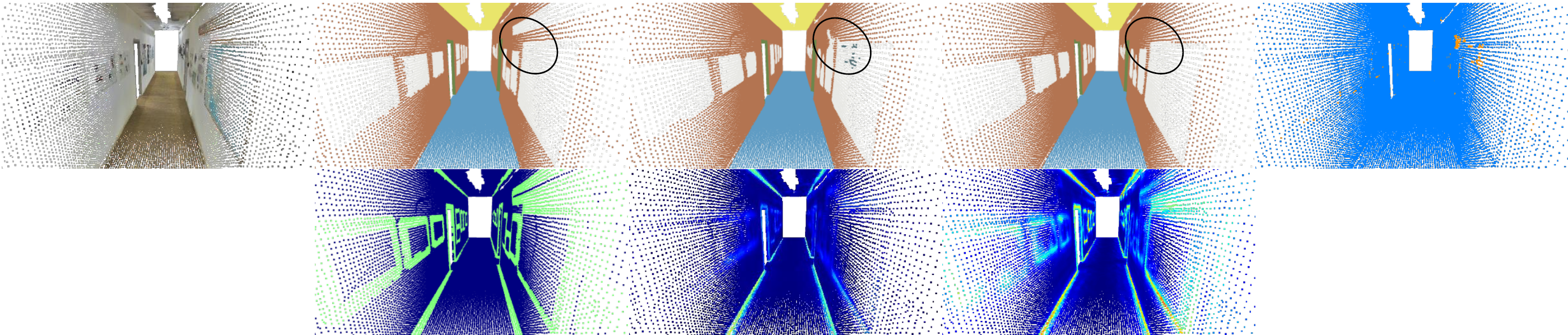}
  \caption{}
  \end{subfigure}
  \begin{subfigure}{\linewidth}
    \includegraphics[width=\linewidth]{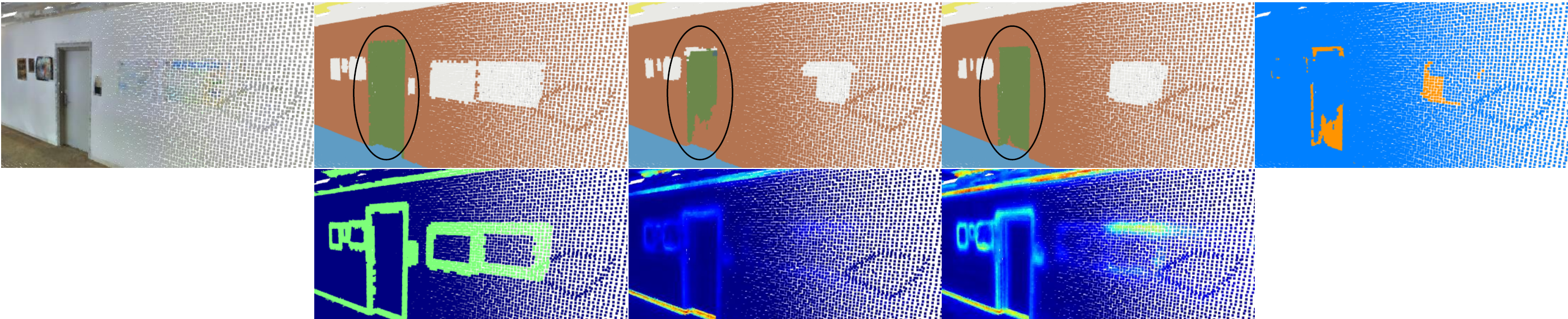}
  \caption{}
  \end{subfigure}
\caption{Hallways. Same as above (\cref{fig:demo_more_room}).}
\label{fig:demo_more_hall}
\end{figure*}

\begin{figure*}[t]  
  \begin{center}
  \begin{subfigure}{\linewidth}
    \includegraphics[width=\linewidth]{plot_more/demo_title.pdf}
    \includegraphics[width=\linewidth]{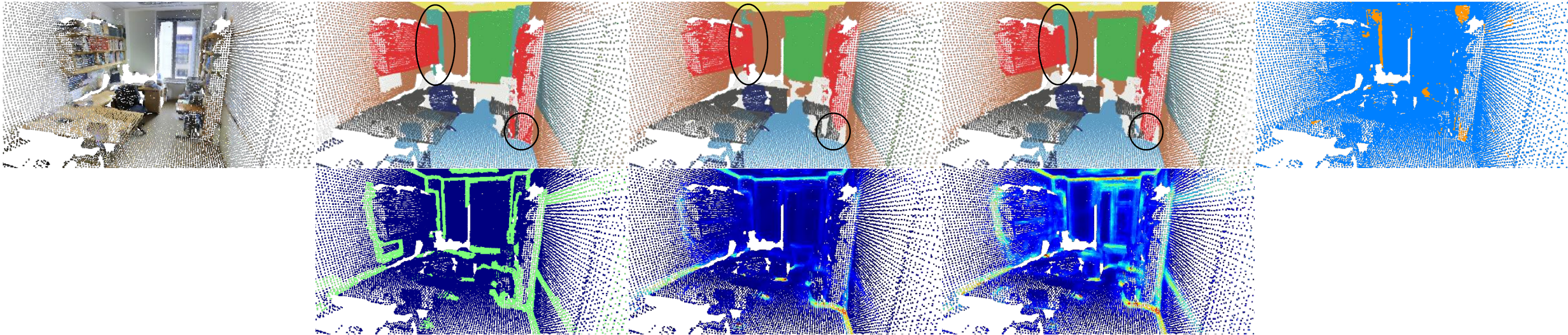}
  \caption{}
  \end{subfigure}

  \begin{subfigure}{\linewidth}
    \includegraphics[width=\linewidth]{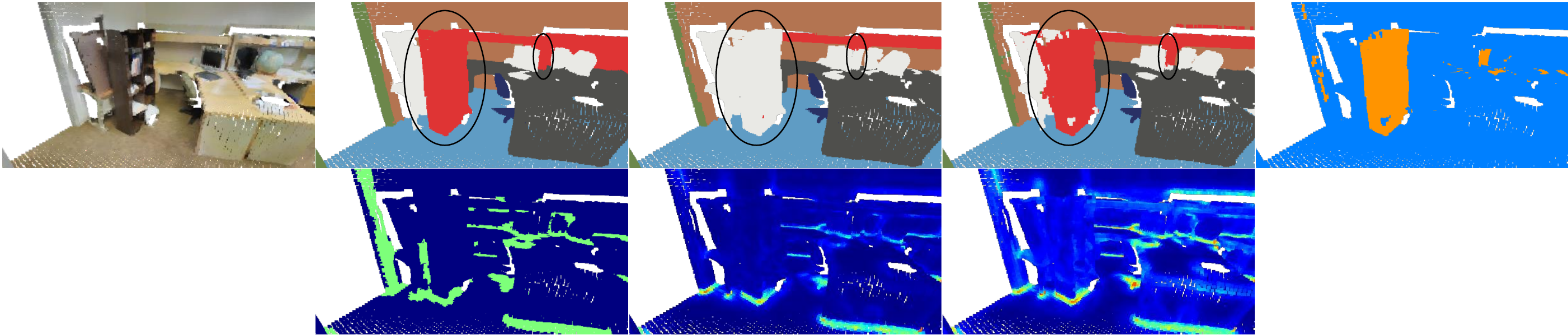}
  \caption{}
  \end{subfigure}
  \begin{subfigure}{\linewidth}
    \includegraphics[width=\linewidth]{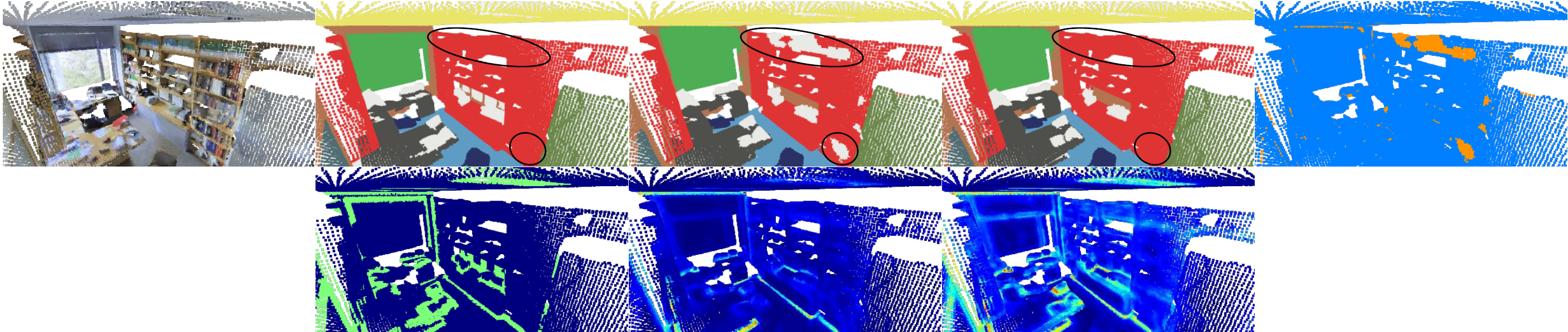}
  \caption{}
  \end{subfigure}
  \end{center}
  \caption{Offices. Same as above (\cref{fig:demo_more_room}).}
  \label{fig:demo_more_office}
\end{figure*}

We provide more qualitative results as a support for the improvement made by CBL on boundaries.
The visualization results include various scenes, including rooms (\cref{fig:demo_more_room}), cluttered space (\cref{fig:demo_more_clutter}), hallways (\cref{fig:demo_more_hall}), and offices (\cref{fig:demo_more_office}). For each scene, we further attempt to visualize the features discrimination between center points and their corresponding neighbors and the results are presented in the every second row. Specifically, we calculate the normalized feature distance between the point feature $f_i$ and features of its neighboring points $\{ f_j ~|~ x_j\in\mathcal N_i\}$. We then take the mean distance for visualization.

According to the presented figures, it shows that the CBL significantly enhances the feature distances around the scene boundaries and improves the baseline to obtain a more detailed and cleaner boundary in prediction for different type of scenes. The visualization is done on S3DIS testset Area 5.

\begin{table}
\centering
\resizebox{\linewidth}{!}{%

\begin{tabular}{r|cc|cc|cc}
\hline
                              & \multicolumn{2}{c|}{mIoU}                & \multicolumn{2}{c|}{OA}                  & \multicolumn{2}{c}{mACC}                 \\
\multirow{-2}{*}{baselines ( + CBL)}     & \cellcolor[HTML]{ECF4FF}boundary & inner & \cellcolor[HTML]{ECF4FF}boundary & inner & \cellcolor[HTML]{ECF4FF}boundary & inner \\ \hline
RandLA-Net\cite{randlanet}    & \cellcolor[HTML]{ECF4FF}+3.3     & +1.4  & \cellcolor[HTML]{ECF4FF}+4.1     & -0.3  & \cellcolor[HTML]{ECF4FF}+3.4     & +2.4  \\
CloserLook3D\cite{closerlook} & \cellcolor[HTML]{ECF4FF}+0.6     & +0.2  & \cellcolor[HTML]{ECF4FF}+0.1     & +0.2  & \cellcolor[HTML]{ECF4FF}+0.7     & +0.4  \\
ConvNet                       & \cellcolor[HTML]{ECF4FF}+2.5     & +2.0  & \cellcolor[HTML]{ECF4FF}+1.0     & +0.7  & \cellcolor[HTML]{ECF4FF}+3.2     & +2.8  \\ \hline
\end{tabular}

}%
\caption{The improvement brought by CBL on different baselines and types of area (boundary / inner area).}
\label{tbl:improve_bound_inner}
\end{table}

\begin{table*}%
\begin{subtable}{\linewidth}%
\begin{center}%
\resizebox{\linewidth}{!}{%
\begin{tabular}{r |c c c | c c c c c c c c c c c c c}%
\hline%
methods                         & mIoU  & OA  & mACC  & ceiling & floor & wall & beam & column & window & door & table & chair & sofa & bookcase & board & clutter \\%
\hline%
RandLA-Net\cite{randlanet}      & 44.1 & 67.1 & 59.1 & 65.5 & 69.4 & 52.2 & 0.0 &  21.4 & 28.6 & 55.0 & 55.0 & 56.0 & 41.1 & 41.2 & 45.8 & 42.1 \\%
\textbf{+ CBL}                  & {\color{red}47.4} & {\color{red}71.2} & {\color{red}62.5} & {\color{red}78.2} & {\color{red}85.9} & {\color{red}56.0} & 0.0 &  {\color{red}30.3} & 25.7 & 42.6 & {\color{red}58.4} & {\color{red}60.9} & {\color{red}50.0} & {\color{red}42.5} & {\color{red}52.2} & {\color{red}44.2} \\%
%
\hline%
CloserLook3D\cite{closerlook}   & 50.0 & 76.6 & 58.5 & 80.7 & 88.6 & 63.9 & 0.0 &  21.1 & 15.6 & 57.5 & 73.3 & 64.7 & 52.2 & 43.1 & 37.2 & 52.6 \\%
\textbf{+ CBL}                  & {\color{red}50.6} & {\color{red}76.7} & {\color{red}59.2} & {\color{red}80.9} & 88.6 & {\color{red}64.6} & 0.0 &  {\color{red}26.5} & 15.6 & 55.9 & 73.0 & {\color{red}65.0} & 50.4 & {\color{red}47.6} & {\color{red}38.4} & 51.2 \\%
\hline%
ConvNet                         & 50.1 & 76.5 & 58.3 & 80.4 & 88.3 & 63.5 & 0.0 &  26.5 & 15.2 & 58.3 & 72.1 & 63.4 & 52.3 & 40.8 & 38.7 & 52.2 \\%
\textbf{+ CBL}                  & {\color{red} 52.6} & {\color{red} 77.5} & {\color{red} 61.5} & {\color{red} 80.5} & {\color{red} 88.8} & {\color{red} 65.7} & 0.0 &  {\color{red}32.5} & {\color{red}20.9} & {\color{red}61.8} & 71.7 & 62.4 & {\color{red}52.5} & {\color{red}46.7} & {\color{red}47.4} & {\color{red}52.5} \\%
\hline%
\end{tabular}%
}%
\end{center}%
\caption{%
The full metrics calculated on boundary points from ground truth (\ie, $\mathcal B_l$) only.%
}%
\label{tbl:bound_iou_l}%
\end{subtable}%
\par\medskip%
\begin{subtable}{\linewidth}
    \begin{center}
    \resizebox{\linewidth}{!}{%
    \begin{tabular}{r |c c c | c c c c c c c c c c c c c}
    \hline
    methods & mIoU  & OA    & mACC  & ceiling  & floor & wall & beam & column & window & door & table & chair & sofa & bookcase & board & clutter     \\
    
    \hline
    RandLA-Net\cite{randlanet}     & 65.8 & 89.6 & 73.0 & 93.3 & 98.6 & 84.6 & 0.0 & 25.9 & 65.7 & 46.5 & 81.1 & 88.9 & 65.4 & 75.5 & 71.9 & 58.2 \\
    \textbf{+ CBL}                 & {\color{red}67.2} & 89.3 & {\color{red}75.4} & 93.0 & 99.1 & 84.6 & 0.0 & {\color{red}37.3} & 64.1 & 39.4 & {\color{red}82.7} & {\color{red}91.5} & {\color{red}79.3} & {\color{red}75.9} & {\color{red}73.9} & 56.0 \\

    \hline
    CloserLook3D\cite{closerlook}  & 70.7 & 92.2 & 75.2 & 96.4 & 99.9 & 86.5 & 0.0 & 25.9 & 55.1 & 76.5 & 95.9 & 87.1 & 81.9 & 75.1 & 72.5 & 66.2 \\
    \textbf{+ CBL}                 & {\color{red}70.9} & {\color{red}92.4} & {\color{red}75.6} & {\color{red}96.5} & 99.9 & {\color{red}86.9} & 0.0 & {\color{red}27.0} & {\color{red}59.3} & {\color{red}78.1} & 95.7 & {\color{red}87.7} & 80.8 & {\color{red}75.4} & 69.4 & 65.6 \\
    
    \hline
    ConvNet                        & 71.2 & 92.1 & 75.5 & 95.0 & 99.8 & 85.9 & 0.0 & 34.6 & 56.0 & 82.7 & 95.4 & 87.4 & 81.3 & 73.8 & 68.4 & 65.7 \\
    \textbf{+ CBL}                 & {\color{red}73.2} & {\color{red}92.8} & {\color{red}78.3} & {\color{red}95.3} & {\color{red}99.9} & {\color{red}88.0} & 0.0 & {\color{red}38.4} & {\color{red}62.2} & 76.4 & {\color{red}95.9} & {\color{red}87.5} & {\color{red}82.7} & {\color{red}81.2} & {\color{red}75.2} & {\color{red}68.6} \\

    \hline
    \end{tabular}
    }%
    \end{center}
    \caption{
    The full metrics calculated on inner points from ground truth (\ie, $\mathcal X - \mathcal B_p$) only.
    }
    \label{tbl:inner_iou_l}
\end{subtable}

\caption{The improvement CBL brought on baselines, separately calculated in boundary area (a) and inner area (b). The {\color{red} red} denotes improvement is made on baseline.}
\label{tbl:bound_inner_iou}

\end{table*}

\begin{table}
\centering
\begin{tabular}{c|ccc}
\hline
temperature & mIoU  & OA    & mACC  \\ \hline
0.3         & 70.67 & 89.16 & 77.91 \\
0.5         & 70.98 & 89.31 & 78.27 \\
1           & 71.33 & 89.40 & 78.69 \\
2           & 70.73 & 89.10 & 77.98 \\
10          & 70.03 & 88.97 & 77.58 \\ \hline
\end{tabular}
\vspace{3pt}
\caption{The effect of temperature on CBL.}
\label{tbl:ablation:temperature}
\vspace{-10pt}
\end{table}

\begin{table*}
\begin{center}
\resizebox{\linewidth}{!}{%
\begin{tabular}{r |c|ccccccccccc }
\hline
    & mIoU (\%) &	Ground &	Building &	Pole &	Bollard &	Trash can &	Barrier &	Pedestrian &	Car &	Natural \\
\hline
HDGCN\cite{hdgcn}	            & 68.3 &	99.4 &	93.0 &	67.7 &	75.7 &	25.7 &	44.7 &	37.1 &	81.9 &	89.6 \\
ConvPoint\cite{convpoints}	    & 75.9 &	99.5 &	95.1 &	71.6 &	88.7 &	46.7 &	52.9 &	53.5 &	89.4 &	85.4 \\
RandLANet\cite{randlanet}	    & 78.5 &	99.5 &	97.0 &	71.0 &	86.7 &	50.5 &	65.5 &	49.1 &	95.3 &	91.7 \\
KP-Conv\cite{kpconv}	        & 82.0 &	99.5 &	94.0 &	71.3 &	83.1 &	78.7 &	47.7 &	78.2 &	94.4 &	91.4 \\
FKAConv\cite{fkaconv}	        & 82.7 &	99.6 &	98.1 &	77.2 &	91.1 &	64.7 &	66.5 &	58.1 &	95.6 &	93.9 \\
PyramidPoint\cite{pyramidpoint} & 82.9 &	99.6 &	97.1 &	74.6 &	84.3 &	56.0 &	65.9 &	79.1 &	95.1 &	93.9 \\
\hline
ConvNet         & 76.2 &    99.5 &  96.3 &  68.5 &  67.4 &  41.4 &  41.5 &  80.6 &  96.3 &  94.1 \\
\textbf{ + CBL} & {\color{red}78.6} &	99.5 &	{\color{red}96.7} &	{\color{red}72.1} &	{\color{red}72.6} &	{\color{red}46.2} &	{\color{red}60.4} &	70.1 &	{\color{red}97.2} &	93.2 \\
\hline
\end{tabular}
}%
\end{center}
\caption{
Quantitative results on Paris-Lille-3D of NPM3D~\cite{npm3d} benchmark, results obtained from online benchmark site by the time of submission. The red denotes the improvement made on baseline.
}
\label{tbl:npm3d_detail}
\end{table*}

\section{Training Setup in Details}
\label{sec:sup:train}
For the RandLA-Net\cite{randlanet} and CloserLook3D\cite{closerlook} baselines, we follow their instructions of released code for training and evaluation, which are \href{https://github.com/QingyongHu/RandLA-Net}{here} (RandLA-Net) and \href{https://github.com/zeliu98/CloserLook3D}{here} (CloserLook3D), respectively. Especially, in CloserLook3D\cite{closerlook}, there are two non-parametric module, we use the one with sin/cos spatial embedding.

For the ConvNet baseline, we use the SGD optimizer to train for 600 epoch, with a weight decay of $0.001$. We set the initial learning rate to $0.01$ and use a momentum of $0.98$ with a decay rate of $0.1^{1/200}$. It roughly takes 24 hours to train on 4 Nividia v100 GPUs, and we does not observe obvious increase in training time after applying the CBL.

\section{Effect of Temperature in CBL}
\label{sec:sup:ablation:temperature}
We conduct empirical study on ScanNet\cite{scannet} validation set to analyze the effect of temperature $\tau$ in the CBL (\cref{eq:cbl}). We use the ConvNet baseline and train for 600 epoch on training set. As shown in \cref{tbl:ablation:temperature}, we find that
the proper temperature for CBL is within $(0.5, 2)$, and we set the temperature to $\tau=1$ by default.

\section{Effect of Design of Sub-scene annotation}
\label{sec:sup:argmax}
While the sub-scene annotation is a distribution, we only use the simple $\arg\max$ when evaluating the boundary points. Therefore, it raises two particular question: 1) is it necessary to maintain the distribution? 2) is there any better way in utilizing the sub-scene annotation than the $\arg\max$?

In this section, we explore other alternatives and answer to this two questions with a particular focus of how they affect the model performance on boundaries.

\begin{figure}
\centering
\resizebox{\linewidth}{!}{\includegraphics[]{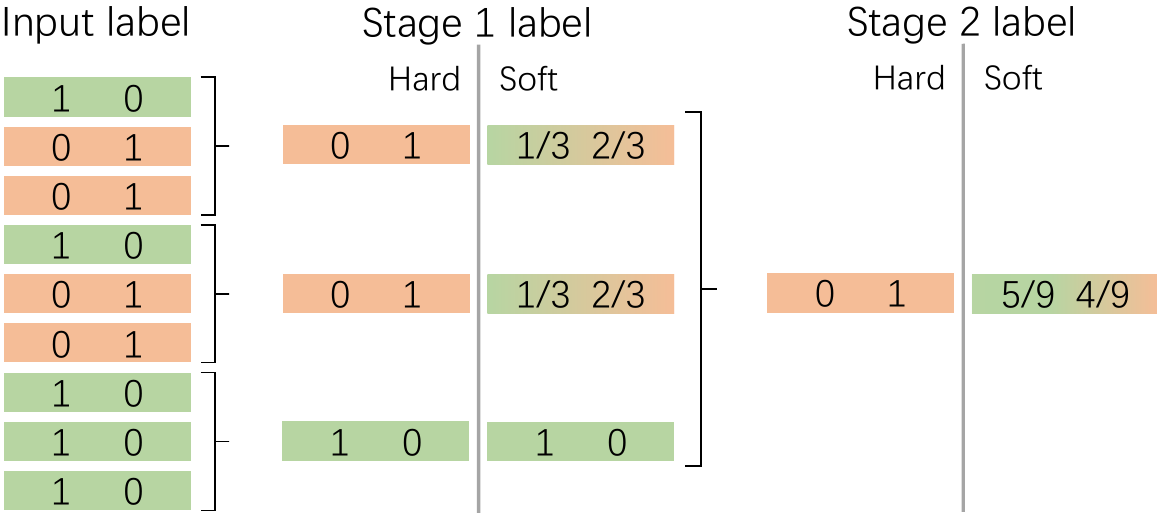}}%
\caption{With every 3 points being sub-sampled into 1 in each stage, tracking distribution (soft label) describes original input faithfully, but hard label fails due to accumulated errors.}
\label{fig:softvshard}
\end{figure}

\noindent\textbf{Necessities of maintaining distribution.}
There are two main reasons to leverage the average pooling on labels and maintain the distribution. First, current methods may not preserve the original input points after sub-sampling, \eg grid sub-sampling in KPConv\cite{kpconv}. Therefore, the original label of a sub-sampled point is not presented and the sub-scene annotation is thus demanded. Although we may use the label of the nearest point for approximation, \cref{tbl:bound_more} shows that CBL (nearest) is sub-optimal. Second, despite that we only use the ``argmax" result of the sub-scene annotation, maintaining distribution still preserves more information than just maintaining ``argmax" result. As ``argmax" discards the minor classes during sampling, such elimination of minority may further accumulate through more sub-sampling stages and leads to imprecise boundary, as depicted in \cref{fig:softvshard}. Experimentally, in \cref{tbl:bound_more}, though CBL (argmax) improves boundary (B-IoU), it compromises overall performance.

\noindent\textbf{Better treatment than Argmax.}
While ``argmax" is straight forward, it introduces the problem of "label-flipping" when the distribution of sub-scene annotation is close to a uniform distribution, \ie, when the number of points of different classes are roughly the same.

To avoid this, we leverage the KL divergence as a measure of the semantic distance among sub-scene annotations. We then threshold on the KL-distance to determine if two sub-scene annotations belong to the same semantic class or not, which further enables us to determine the boundary points in sub-sampled point cloud. Specifically, we set the threhold to $0.5$ and CBL (kl) can be bring a small improvement on overall performance, and a slightly larger boost on boundary performance, as in \cref{tbl:bound_more}. Yet, as ``thresholding KL distance" introduces extra hyper-parameters and complexity, we opt for ``argmax" for simplicity in the main paper.

\noindent\textbf{Summary.}
Therefore, we summarize the reason for designing the sub-scene annotation as a distribution as it can preserve much more information and can be extended to a more robust boundary determination using KL-distance.

\begin{table}[t]
\begin{center}
\centering
\resizebox{\linewidth}{!}{%
\begin{tabular}{l|ccc|c}
\hline
                                     & \multicolumn{3}{c|}{mIoU}                                                                   &                              \\
\multirow{-2}{*}{methods}            & overall                      & @boundary                    & @inner                       & \multirow{-2}{*}{B-IoU}      \\ \hline
ConvNet & 67.4                         & 50.1                         & 71.2                         & 59.6                         \\
\hline
ConvNet + CBL     & 69.4                         & 52.6                         & 73.1                         & 61.5                         \\
ConvNet + CBL (nearest) & 68.3                         & 52.1                   & 71.8                         & 60.9                         \\
ConvNet + CBL (argmax)                 & 66.8                         & 50.6                         & 70.4                         & 60.6                        \\
ConvNet + CBL (kl)               & 69.5                         & 52.5                         & 73.2                         & 62.0                         \\
\hline
\end{tabular}
}%
\end{center}
\caption{Same setting as in \cref{tbl:bound} in main paper.
}
\label{tbl:bound_more}
\end{table}

\section{Further Experiments}
\label{sec:sup:rst}
\noindent\textbf{Results on ScanNet and NPM3D datasets.}
We provide the detail results on ScanNet in \cref{tbl:scannet_detail}; and the detail results on NPM3D in \cref{tbl:npm3d_detail}.

\noindent\textbf{CBL with Transformer.}
We use the open-source code base (\href{https://github.com/POSTECH-CVLab/point-transformer}{here}) to re-produce the performance of newly released point Transformer\cite{pttransformer} on S3DIS\cite{s3dis} Area 5 dataset.

In \cref{tbl:s3dis:pttrans}, the same consistent improvement is made on classes such as column. CBL with better boundaries further boosts the overall performance to 71.0 in mIoU, achieving a new state-of-the-art performance.

\begin{table*}
\begin{center}
\resizebox{\linewidth}{!}{%
\begin{tabular}{r |c| cccccccccccccccccccc }
    \hline
    Method & mIoU &	bathtub &	bed &	books. &	cabinet &	chair &	counter &	curtain &	desk &	door &	floor &	other &	pic &	fridge  &	shower &	sink &	sofa &	table &	toilet &	wall &	wndw \\
    
    \hline
    DCM-Net\cite{dcmnet}    &   65.8 &	77.8 &	70.2 &	80.6 &	61.9 &	81.3 &	46.8 &	69.3 &	49.4 &	52.4 &	94.1 &	44.9 &	29.8 &	51.0 &	82.1 &	67.5 &	72.7 &	56.8 &	82.6 &	80.3 &	63.7 \\
    VMNet\cite{vmnet}       &   74.6 &	87.0 &	83.8 &	85.8 &	72.9 &	85.0 &	50.1 &	87.4 &	58.7 &	65.8 &	95.6 &	56.4 &	29.9 &	76.5 &	90.0 &	71.6 &	81.2 &	63.1 &	93.9 &	85.8 &	70.9 \\
    
    \hline
    SparseConvNet\cite{seg_vx_SSCN}  &  72.5 &	64.7 &	82.1 &	84.6 &	72.1 &	86.9 &	53.3 &	75.4 &	60.3 &	61.4 &	95.5 &	57.2 &	32.5 &	71.0 &	87.0 &	72.4 &	82.3 &	62.8 &	93.4 &	86.5 &	68.3 \\
    MinkowskiNet\cite{Minkowski}     &  73.6 &	85.9 &	81.8 &	83.2 &	70.9 &	84.0 &	52.1 &	85.3 &	66.0 &	64.3 &	95.1 &	54.4 &	28.6 &	73.1 &	89.3 &	67.5 &	77.2 &	68.3 &	87.4 &	85.2 &	72.7 \\
    O-CNN\cite{ocnn}                 &  76.4 &	75.8 &	79.6 &	83.9 &	74.6 &	90.7 &	56.2 &	85.0 &	68.0 &	67.2 &	97.8 &	61.0 &	33.5 &	77.7 &	81.9 &	84.7 &	83.0 &	69.1 &	97.2 &	88.5 &	72.7 \\
    OccuSeg\cite{occuseg}            &  76.2 &	92.4 &	82.3 &	84.4 &	77.0 &	85.2 &	57.7 &	84.7 &	71.1 &	64.0 &	95.8 &	59.2 &	21.7 &	76.2 &	88.8 &	75.8 &	81.3 &	72.6 &	93.2 &	86.8 &	74.4 \\
    Mix3D\cite{mix3d}                &  78.1 &	96.4 &	85.5 &	84.3 &	78.1 &	85.8 &	57.5 &	83.1 &	68.5 &	71.4 &	97.9 &	59.4 &	31.0 &	80.1 &	89.2 &	84.1 &	81.9 &	72.3 &	94.0 &	88.7 &	72.5 \\
    
    \hline
    BA-GEM\cite{bound_3d_pred} * &  63.5 &	     &	     &	     &	     &	     &	     &	     &	     &	     &	     &  	 &  	 &  	 &  	 &  	 &  	 &  	 &  	 &  	 &	     \\
    PointConv\cite{pointconv}    &  66.6 &	78.1 &	75.9 &	69.9 &	64.4 &	82.2 &	47.5 &	77.9 &	56.4 &	50.4 &	95.3 &	42.8 &	20.3 &	58.6 &	75.4 &	66.1 &	75.3 &	58.8 &	90.2 &	81.3 &	64.2 \\
    PointASNL\cite{PointASNL}    &  66.6 &	70.3 &	78.1 &	75.1 &	65.5 &	83.0 &	47.1 &	76.9 &	47.4 &	53.7 &	95.1 &	47.5 &	27.9 &	63.5 &	69.8 &	67.5 &	75.1 &	55.3 &	81.6 &	80.6 &	70.3 \\
    KP-Conv\cite{kpconv}         &  68.4 &	84.7 &	75.8 &	78.4 &	64.7 &	81.4 &	47.3 &	77.2 &	60.5 &	59.4 &	93.5 &	45.0 &	18.1 &	58.7 &	80.5 &	69.0 &	78.5 &	61.4 &	88.2 &	81.9 &	63.2 \\
    FusionNet\cite{fusionnet}    &  68.8 &	70.4 &	74.1 &	75.4 &	65.6 &	82.9 &	50.1 &	74.1 &	60.9 &	54.8 &	95.0 &	52.2 &	37.1 &	63.3 &	75.6 &	71.5 &	77.1 &	62.3 &	86.1 &	81.4 &	65.8 \\
    JSENet\cite{bound_3d_jse}    &  69.9 &	88.1 &	76.2 &	82.1 &	66.7 &	80.0 &	52.2 &	79.2 &	61.3 &	60.7 &	93.5 &	49.2 &	20.5 &	57.6 &	85.3 &	69.1 &	75.8 &	65.2 &	87.2 &	82.8 &	64.9 \\
    RFCR\cite{omni}              &  70.2 &	88.9 &	74.5 &	81.3 &	67.2 &	81.8 &	49.3 &	81.5 &	62.3 &	61.0 &	94.7 &	47.0 &	24.9 &	59.4 &	84.8 &	70.5 &	77.9 &	64.6 &	89.2 &	82.3 &	61.1 \\
    
    \hline
    \textbf{ConvNet + CBL}  & 70.5 & 76.9 & 77.5 & 80.9 & 68.7 & 82.0 & 43.9 & 81.2 & 66.1 & 59.1 & 94.5 & 51.5 & 17.1 & 63.3 & 85.6 & 72.0 & 79.6 & 66.8 & 88.9 & 84.7 & 68.9 \\
    \hline
\end{tabular}
}%
\end{center}
\caption{
Quantitative results on ScanNet~\cite{scannet} benchmark, results obtained from online benchmark site by the time of submission.
We group method by the 3D representation type, which is respectively, from top to down, 3D + mesh, 3D voxel and 3D point, and we also use 3D point. The empty line denotes no record of detailed performance found. The method with * also considers boundary.
}
\label{tbl:scannet_detail}
\end{table*}

\begin{table*}[h!]
    \begin{center}
    \resizebox{\linewidth}{!}{%
    \begin{tabular}{r|lll|lllllllllllll}
    \hline
    methods & mIoU & OA & mACC & ceiling & floor & wall & beam & column & window & door & table & chair & sofa & bookcase & board & clutter \\
    \hline
        pt trans\cite{pttransformer}* &  70.4 & \hspace{-3pt} 90.8 & \hspace{3pt} 76.5 & \hspace{1pt} 94.0 &  98.5 & \hspace{-1pt} 86.3 & \hspace{1pt} 0.0 & \hspace{3pt} 38.0 & \hspace{4pt} 63.4 & \hspace{-2pt} 74.3 &  89.1 & \hspace{-1pt} 82.4 & \hspace{-2pt} 74.3 & \hspace{6pt} 80.2 &  76.0 &  59.3 \\
        \hline
        pt trans\cite{pttransformer} &  70.0 & \hspace{-3pt} 90.5 & \hspace{3pt} 76.5 & \hspace{1pt} 95.2 &  98.6 & \hspace{-1pt} 85.1 & \hspace{1pt} 0.0 & \hspace{3pt} 36.7 & \hspace{4pt} 62.5 & \hspace{-2pt} 75.9 &  81.5 & \hspace{-1pt} 91.0 & \hspace{-2pt} 75.1 & \hspace{6pt} 71.9 &  76.4 &  60.2
        \\
        \textbf{ + CBL}               &  {\color{red}71.0*} & \hspace{-3pt} {\color{red}90.9*} & \hspace{3pt} {\color{red}77.5*} & \hspace{1pt} 94.3* &  98.3 & \hspace{-1pt} {\color{red}87.4*} & \hspace{1pt} 0.0 & \hspace{3pt} {\color{red}42.1*} & \hspace{4pt} {\color{red}64.0*} & \hspace{-2pt} {\color{red}78.5*} &  {\color{red}82.5} & \hspace{-1pt} 88.9* & \hspace{-2pt} 75.1* & \hspace{6pt} 71.1 &  {\color{red}81.3*} &  59.6*
        \\
        \hline
    \end{tabular}%
    }%
\end{center}%
\caption{%
Quantitative results on S3DIS Area 5 dataset~\cite{s3dis}, showing the mean IoU (mIoU), overall accuracy (OA), mean accuracy (mACC), and per-class IoU scores. We include both performance reported in original paper (with *, the first row) and the re-produced performance (without *, the second row). We use {\color{red}red} to denote improvement over the re-produced point transformer, and * to denote the improvement over the performance reported in original paper.
}
\label{tbl:s3dis:pttrans}
\end{table*}

\end{document}